\documentclass{article}[12pt, a4]
\usepackage[greek,english]{babel}
\usepackage{amssymb}
\usepackage{amsthm}
\usepackage{amsmath}
\usepackage{enumitem}
\usepackage{hyperref}
\usepackage{authblk}
\usepackage{mathtools}

\usepackage{subcaption}

\newtheorem{theorem}{\large Theorem}[section]

\newtheorem{definition}{\large Definition}[section]

\newtheorem{example}{\large Example}[section]

\newtheorem{proposition}{\large Proposition}[section]

\def\1{\rule{0pt}{1.7ex}xy}

\def\2{\rule{0pt}{1.7ex}x_0x}
\def\3{\rule{0pt}{2ex}X_x}

\def\4{\rule{0pt}{1.7ex}1}
\def\5{\rule{0pt}{1.7ex}2}

\begin{document}
	
	\thispagestyle{empty}
	
	\title{\Large Neural Network Operator-Based Fractal Approximation: Smoothness Preservation and Convergence Analysis}
	
	\author[a]{Aaqib Ayoub Bhat}
	
	\author[a]{Asif Khan}

	\author[a,b,c,*]{M Mursaleen}

\affil[a]{\it \small Department of Mathematics, Aligarh Muslim University, Aligarh, 202002, Uttar Pradesh, India}

\affil[b]{\it \small Department of Medical Research, China Medical University Hospital, China Medical University (Taiwan), Taichung, Taiwan}

\affil[c]{\it \small Faculty of Mathematics and Natural Sciences, Universitas Sumatera Utara, Medan 20155, Indonesia}

\date{}

\maketitle

\let\thefootnote\relax\footnotetext{*corresponding author (M Mursaleen)\\ Email addresses:\\ bhataqib19@gmail.com (Aaqib Ayoub Bhat),\\ akhan.mm@amu.ac.in (Asif Khan), \\mursaleenm@gmail.com (M Mursaleen)}

\begin{abstract}
	This paper introduces the construction of fractal interpolation functions (FIFs), whose graphs are the attractors of an iterated function system (IFS). Integrating concepts from approximation theory, $\alpha$-fractal functions are constructed, employing shallow neural network operators. Based on the same methodology, we developed fractal interpolation functions using only discrete function values, unlike traditional methods that require each value of the target function. In order to preserve the smoothness of the target function, a method for constructing such FIFs is introduced, employing four-layered neural network operators i.e., whenever $f \in C^{r}[a,b]$, the corresponding FIF $f^{\alpha} \in C^{r}[a,b]$. This work uses key approximation theory tools, such as the modulus of continuity and interpolation operators, to develop convergence results and uniform approximation error bounds. To validate the theoretical results obtained, numerical experiments with graphical analysis using Python is provided.

\end{abstract}


\maketitle

\section{Introduction}\label{sec1}

Approximation theory is a fundamental branch of mathematics with applications in computer science, engineering, etc. It is considered an important tool in data processing, modeling, and signal processing. The study of interpolation and approximation has a long and rich history. Traditional methods, such as Taylor series approximation, B-splines, Lagrange interpolation, function approximation \cite{ aymanfilomat, cai2023approximation, cai2018shape}, etc., have been widely explored and applied in various fields. With the advancement of approximation theory, the concept of approximation was extended to approximation by operators e.g., \cite{articlekj, khan2021phillips, MURSALEEN2015874, far} However, these methods proved less effective in the cases where the data is uneven or has self-similar patterns in some sense. Barnsley \cite{barnsley1986fractal} in the 1980s led to the development of fractal approximation by developing fractal interpolation functions (FIFs) whose graphs are the attractors of iterated function systems. By utilizing the geometric properties of fractals, FIFs offer a highly adaptable framework for data interpolation. This approach proves effective in interpolating and approximating highly irregular datasets, and thus bridging the gap between mathematical theory and practical applications.

In the recent past, neural networks have been used in the task of modeling and approximation of various structures in different areas such as image processing, time series analysis, machine learning, and many others. Integrating neural networks with fractal-based approaches opens new ways of developing effective interpolation techniques. This paper introduces the idea of neural network \(\alpha\)-fractal interpolation functions in which we combine the neural network operators with the capabilities of fractals to interpolate and approximate on the set of all continuous functions.

A fundamental requirement for the analysis of non-smooth datasets is the quantification of their geometric irregularity. The study of fractal dimension is the quantitative measure of the geometric irregularity, complexity and self-similarity of a set, which expands the classical notion of topological dimension to non-integer values (Mandelbrot, 1967 \cite{doi:10.1126/science.156.3775.636}). Natural forms like coastlines, borders and terrain do not always adapt to the smooth, differentiable curves assumed by classical methods, so fractal-based techniques are essential to precisely capture them. This can be achieved using fractal interpolation functions (FIFs) that are built from iterated function systems and produce self-affine curves and whose roughness can be tuned, typically via vertical scaling (or contractivity) factors, to that of the fractal dimension of the underlying data. Thus, FIFs can be used to approximate irregular real world datasets and to estimate the fractal dimension of their geometry. Nasim et al. \cite{nasim} determined the box dimension of $\alpha$-fractal interpolation functions by imposing suitable conditions on the original, base, and scaling functions. Hussain et al \cite{husain2024novel} developed an algorithm based on FIF for estimation of fractal dimension and reconstruction of coastline geometry, building on their earlier work on the fractal dimension of the Australian coastline \cite{husain2021fractal} and the box dimension of the Saudi Arabian border \cite{sajid2023box}.

Over time, fractal functions have been explored for their approximation properties within various function spaces, in particular the space of continuous functions. Massopust \cite{MASSOPUST1997171} worked on the theory of fractal functions and demonstrated the applications in finite elements for obtaining the solutions of partial differential equations.
Navascués et al. \cite{Navascues2005, navascues2010fractal, navascues2006smooth} investigated various aspects of the $\alpha$-fractal function $f^{\alpha}$ of $f$, including its approximation capabilities. Similarly Vijender \cite{vijenderbernstein} introduced fractal interpolation functions based on Bernstein operators.
At the same time, neural networks have emerged as powerful tools for modeling and approximating complex structures. The universal approximation theorem \cite{HORNIK1989359} demonstrated that neural networks are capable of approximating any continuous function on a compact domain to any preassigned degree of accuracy. Later many authors introduced numerous neural network interpolation and approximation operators  \cite{BHAT2026190, bhat2026blending, Brewal2026, COSTARELLI201528,  COSTARELLI201480}. This has inspired a fusion of neural networks with fractal-based approaches, resulting in innovative hybrid methods for interpolation and approximation.

The following is a brief introduction to sigmoidal functions, which are used as activation functions in neural network operators and will be utilized throughout this article

\begin{definition}
	A measurable function $\rho :\mathbb{R}\rightarrow \mathbb{R}$ is said to be sigmoidal  function if $\rho$ satisfies the following:\\
	
	$\displaystyle\lim_{x\rightarrow \infty}\rho(x)=1,$ and $\displaystyle\lim_{x\rightarrow -\infty}\rho(x)=0.$ 
\end{definition}

\noindent Qian et al. \cite{QIAN2022126781} introduced a newly class of sigmodal functions $\mathcal{A}(m)$ defined as follows:\\

\begin{definition}\label{sig}
	For fixed $m\in \mathbb{R}^+$, we say a sigmoidal function $\rho \in \mathcal{A}(m)$ if it satisfies the following two conditions:\\
	\begin{enumerate}
		\item[(i)] $\rho(x)$ is nondecreasing;
		\item[(ii)] $\rho(x)=1$ for $x\geq m$, and $\rho(x)=0$ for $x\leq -m.$
	\end{enumerate}
\end{definition}

Now, define a new function $\xi$ as follows:
\begin{equation}\label{xi}
	\xi(x)=\rho(x+m)-\rho(x-m).
\end{equation}
The function $\xi$ serves as the activation function in the hidden layer of neural network operators used in this paper.

\section{Fractal Interpolation}\label{sec2}
In this article, we use the following notations and terminology. The set of real numbers is denoted by $\mathbb{R}$, and the set of natural numbers is written as $\mathbb{N}$. For any $N \in \mathbb{N}$, the first $N$ natural numbers are represented by $\mathbb{N}_N$. The space $\mathcal{C}^{r}[a, b]$ refers to the collection of all real-valued functions on the interval $[a, b]$ that are $r$-times differentiable, with the $r$-th derivative being continuous. Throughout this article FIF will refer to fractal interpolation function.
	
	Let \( a=x_0 < x_1 < x_2 < \cdots < x_N=b \) for some \( N \in \mathbb{N} \) with \( N > 2 \) define a partition of the closed interval \([a, b]\). Consider a set of real numbers \( y_0, y_1, y_2, \ldots, y_N \). Let \( \pounds_i \), where \( i =1,2,\cdots,N \), represent a collection of homeomorphic mappings from \([a, b]\) to \([x_{i-1}, x_i]\) that satisfy the following properties:
	\begin{equation}\label{ln}
		\pounds_{i}\left(x_{0}\right)=x_{i-1}, \quad \pounds_{i}\left(x_{N}\right)=x_{i}.
	\end{equation}

	Let \( \mathcal{J}_i \), \(\forall ~  i =1,2,\cdots,N\) be the functions defined on \( [a, b] \times \mathbf{C} \) with values in \(\mathbf{C}\), where \(\mathbf{C}\) is a compact subset of \( \mathbb{R} \). The functions \( \mathcal{J}_{i} \) are continuous with respect to \( x \) and contractive with respect to \( y \), with a contraction factor \(|\alpha_i| \leq \kappa < 1 \), such that
	\begin{equation}\label{fn}
		\mathcal{J}_{i}\left(x_{0}, y_{0}\right)=y_{i-1}, \quad \mathcal{J}_{i}\left(x_{N}, y_{N}\right)=y_{i}.
	\end{equation}
	Defining a mapping \(\lambda_{i}: [a,b] \times\mathbf{C}\to [x_{i-1},x_i] \times \mathbf{C}\) by  
	\begin{equation*}
		\lambda_{i}(\tilde{x}, \tilde{y}) = \left(\pounds_i(\tilde{x}), \mathcal{J}_{i}(\tilde{x}, \tilde{y})\right), \quad (\tilde{x}, \tilde{y}) \in [a,b] \times \mathbf{C}, \, i =1,2,\cdots,N, 
	\end{equation*}
	Therefore, 
	\begin{equation}\label{ifs00}
		\Lambda = \left\{[a,b] \times \mathbf{C}, \lambda_{i}(x, y) , i =1,2,\cdots,N\right\}.
	\end{equation}
	defines IFS.
	
	Let \(X_{y_0}^{y_N} = \left\{\phi: [a,b] \to \mathbb{R} \mid \phi\in C[a,b], \phi(x_0) = y_0, \text{ and } \phi(x_N) = y_N\right\}\). Clearly $X_{y_0}^{y_N}$ is a subset of normed space $\left(C[a,b], \|\cdot\|_\infty\right)$. The metric $d$ on $X_{y_0}^{y_N}$ is defined as  
	\[
	d(\phi, \psi) = \|\phi - \psi\|_\infty, \quad \forall ,\phi, \psi \in X_{y_0}^{y_N},
	\]
	where $\|\phi - \psi\|_\infty =\displaystyle\sup_{x \in [a, b]} |\phi(x) - \psi(x)|$. This metric is identical to the one induced by the $\|\cdot\|_\infty$ norm on $\mathcal{C}[a, b]$.
	i.e., 
	\begin{equation*}
		d(\phi, \psi) = \underset{x\in [a,b]}{\max}\left|\phi(x) - \psi(x)\right|,\quad\forall~ \phi , \psi \in X_{y_0}^{y_N}.
	\end{equation*} 
	For $i =1,2,\cdots,N$, define the operator $T$ on $(X_{y_0}^{y_N}, d)$ by
	\begin{equation}\label{rb1}
		(T \phi)(x)=\mathcal{J}_{i}\left(\pounds_{i}^{-1}(x), \phi \circ \pounds_{i}^{-1}(x)\right), \quad x \in [x_{i-1},x_{i}].
	\end{equation}
	The operator defined in (\ref{rb1}) is known as Read-Bajraktarević operator (RB operator). The operator \( T \) is well-defined, which can be verified using the properties of \( \pounds_i \) and the relations described in Equations (\ref{ln}) and (\ref{fn}).\\
	Also, since $\mathcal{J}_i$ is contractive with respective to second argument with contraction factor $|\alpha_{i}|$, we have 
	\begin{equation*}
		d(T \phi, T \psi) \leq|\alpha|_{\infty} d(\phi, \psi),
	\end{equation*}
	where $\displaystyle|\alpha|_{\infty}=\max_{i} \left\{\left|\alpha_{i}\right|\right\}<1$.\\
	Therefore, \(T\) is a contraction mapping on \((X_{y_0}^{y_N}, d)\). Moreover, $X_{y_0}^{y_N}$ is a closed subset of Banach space $C[a,b]$ and hence $(X_{y_0}^{y_N},d)$ is complete. Therefore, by Banach contraction theorem, \(T\) admits a unique fixed point, say \(\phi^{\alpha}\) in \(X_{y_0}^{y_N}\). Since $T$ is defined by functional equation given in  (\ref{rb1}), therefore
	\begin{equation*}
		\phi^{\alpha}(x) = \mathcal{J}_{i}\left(\pounds_i^{-1}(x), \phi^{\alpha} \circ \pounds_i^{-1}(x)\right), \quad x \in [x_{i-1}, x_{i}]. 
	\end{equation*} 
	Further, the interpolation properties of $\phi^{\alpha}$ can be verified by using conditions given in Equations (\ref{ln}) and (\ref{fn}).

	Barnsley in his paper \cite{barnsley1986fractal} proved that the graph \(G(\phi^{\alpha})\) of \(\phi^{\alpha}\) satisfy the following:
	\begin{equation*}
		G\left(\phi^{\alpha}\right)=\bigcup_{i =1,2,\cdots,N} \lambda_{i}\left(G\left(\phi^{\alpha}\right)\right).
	\end{equation*}
	Thus, the function \(\phi^{\alpha}\) is known as the FIF associated with the IFS defined in (\ref{ifs00}).
	
	Barnsley's \cite{barnsley1986fractal} concept of FIF extends interpolation theory by constructing continuous functions that are fractal in nature. Using Iterated Function System (IFS), FIF generates self-similar interpolants, allowing the modeling of complex, irregular shapes while preserving continuity at interpolation points. This approach bridges classical interpolation with fractal geometry. Bransley \cite{barnsley1986fractal} and Navascu$\acute{e}$s \cite{Navascues2005} demonstrated that FIFs can be utilized to construct a family of fractal functions corresponding to any continuous real-valued function \(f\) defined on the closed interval \([a, b]\).

	\section{Neural network $\alpha$-fractal approximation}\label{sec3}
	In this section, we develop a theoretical foundation for the construction of fractal interpolation functions (FIFs) corresponding to any continuous real-valued function \(f\) defined on the closed interval $[a, b]$ using an iterated function system (IFS) employing neural network operators. We also study their convergence properties and obtain uniform error estimates by employing classical tools of approximation theory, in particular, the modulus of continuity.
	
	Consider a real valued continuous function $f:[a,b]\rightarrow \mathbb{R}$. Let $\Delta =\{x_0, x_1, \cdots, x_N\} $ be the partition of $[a,b]$ and let $\{(x_i, f(x_i)), \quad i=0,1, \cdots, N\}$ be the data set.
	We draw our special attention to the following two equations:
	\begin{equation}\label{l2}
		\pounds_i(x)= a_i.x +b_i
	\end{equation}
	and 
	\begin{equation}\label{f2}
		\mathcal{J}_i(x,y)=\alpha_{i}.y+\wp_{i}(x)
	\end{equation}
	where $\pounds_i$ and $\mathcal{J}_i$ satisfy the conditions (\ref{ln}) and (\ref{fn}) respectively $\ \forall i=1,2,\cdots,N$.\\
	
	For all $i=1,2,\cdots,N$, we define 
	\begin{equation}\label{qi}
		\wp_{i}(x)=h(\pounds_i(x))-\alpha_{i}.\ss(x)  
	\end{equation}
	where $h(x)$ is a continuous function that satisfies the interpolation condition
	\begin{align}\label{hi}
		h(x_i)=f(x_i)
	\end{align}
	and $\ss(x)$ is another continuous function that satisfies the boundary conditions
	\begin{align}\label{b1}
		\ss(x_0)=f(x_0) \quad\text{and}\quad \ss(x_N)=f(x_N)
	\end{align}\\
	In this context, we designate $\ss(x)$ as the base function and $h(x)$ as the height function. The function $f(x)$ remains invariant under a sequence of transformations: subtracting $\ss(x)$ from $f(x)$, applying a vertical scaling to the difference by a factor of $\alpha_i$, compressing the result horizontally so that the interval $[a, b]$ is mapped onto $[x_{i-1}, x_i]$, and subsequently adding this adjusted expression to $h(x)$ over the interval $[x_{i-1}, x_i]$ for each $i$.\\
	
	Let $\ss$ be defined by neural network operator $S_{n,\rho}(f, .)$ introduced by Qian et al in \cite{QIAN2022126781} i.e.,
	\begin{align}
		\ss(x)=S_{n,\rho}(f, x)= \sum_{k=0}^{n}f(a_k)\xi\left(\frac{2m}{h}(x-a_k)\right)\quad \forall~ x \in [a,b], \forall ~n \in \mathbb{N} .
	\end{align}
	where $a_k=a+kh$ and $h=\frac{b-a}{n}$.\\

	$S_{n,\rho}(f, .)$ interpolates the function $f$ at each point $a_k,~k=0,1,2,\cdots, n$ as discussed in \cite{QIAN2022126781}. In particular it interpolates the function $f$ at $x_0$ and $x_N$ i.e.,  $S_{n,\rho}\left(f, x_{0}\right)=f\left(x_{0}\right), S_{n,\rho}\left(f, x_{N}\right)=f\left(x_{N}\right)$ for all $n \in \mathbb{N}$.
	Define IFS through following functions:
	\begin{equation}\label{defli}
		\pounds_i(x)= \frac{x_i -x_{i-1}}{x_N-x_0}x +\frac{x_N x_{i-1} -x_0 x_i}{x_N-x_0}, \quad \forall i=1,2,\cdots,N
	\end{equation}
	and 
	\begin{equation}\label{deffi}
		\mathcal{J}_i(x,y)=\alpha_{i}.y+f(\pounds(x))-\alpha_{i}S_{n,\rho}(f, x), \quad \forall i=1,2,\cdots,N.
	\end{equation}\\
	It can be easily shown that $\pounds_i$ satisfies the condition (\ref{ln}). Now we will show that $\mathcal{J}_i$ satisfies the condition (\ref{fn}) as follows:
	\begin{align*}
		\mathcal{J}_i(x_0,f(x_0))&=\alpha_{i}.f(x_0)+f(\pounds(x_0))-\alpha_{i}S_{n,\rho}(f, x_0)\\
		&=\alpha_{i}.f(x_0)+f(x_{i-1})-\alpha_{i}.f(x_0)\\
		&=f(x_{i-1}).
	\end{align*}
	Similarly, one can show that
	\begin{align*}
		\mathcal{J}_i(x_N,f(x_N))=f(x_i)
	\end{align*}
	Also $\mathcal{J}_i$ satisfies the Lipschitz condition in second argument i.e.,
	\begin{align*}
		|\mathcal{J}_i(x,y)-\mathcal{J}_i(x,z)|\leq |\alpha_{i}||y-z|.
	\end{align*}
	Let \begin{align*}
		\lambda_{i}(x,y)=(\pounds_i(x), \mathcal{J}_i(x,y)),\quad \forall i=1,2,\cdots,N.
	\end{align*}
	Define $X_{\beta_1}^{\beta_2}=\left\{\phi\in C[a,b]: \phi(a)=\beta_1 \quad \text{and} \quad \phi(b)=\beta_2\right \}$ for fixed $\beta_1$, $\beta_2$ $\in$ $\mathbb{R}$.\\
	$(X_{\beta_1}^{\beta_2},\|\cdot\|_\infty)$ is a closed subset of $(C[a,b],\|\cdot\|_\infty)$ and hence complete. Choose $\beta_1=f(a)$ and $\beta_2=f(b)$. \\
	Define a mapping $T:X_{\beta_1}^{\beta_2}\rightarrow X_{\beta_1}^{\beta_2}$ by
	\begin{align}\label{cont}
		T\phi(x)=\mathcal{J}_i({\pounds_i}^{-1}(x),\phi\circ {\pounds_i}^{-1}(x)), \quad x \in [x_{i-1}, x_{i}], \quad i =1,2,\cdots,N.
	\end{align}
	$T$ is a contraction:
	\begin{align*}
		|T\phi(x)-T\psi(x)|&=|\mathcal{J}_i({\pounds_i}^{-1}(x),\phi\circ {\pounds_i}^{-1}(x))-\mathcal{J}_i({\pounds_i}^{-1}(x),\psi\circ {\pounds_i}^{-1}(x))|\\
		&\leq |\alpha_{i}||\phi\circ {\pounds_i}^{-1}(x))-\psi\circ {\pounds_i}^{-1}(x))|\\
		&= |\alpha_{i}||(\phi - \psi) ({\pounds_i}^{-1}(x))|\\
		&\leq |\alpha_{i}| \|(\phi - \psi)\|_{\infty}.
	\end{align*}\\
	Now, $T $ is a contraction on $X_{\beta_1}^{\beta_2}$ and $X_{\beta_1}^{\beta_2}$ being complete, hence by Banach Contraction theorem, $T$ admits a unique fixed point say $f_{n}^{\alpha}$ on $X_{\beta_1}^{\beta_2}$. Therefore, Equations (\ref{deffi}) and  (\ref{cont}) gives
	\begin{align}\label{fif}
		f_{n}^{\alpha}(x)&=\mathcal{J}_i({\pounds_i}^{-1}(x),f_{n}^{\alpha}\circ {\pounds_i}^{-1}(x)) \nonumber\\
		&=\alpha_{i}f_{n}^{\alpha}(\pounds_{i}^{-1}(x))+f(x)-\alpha_{i}S_{n,\rho}(f, \pounds_{i}^{-1}(x)),~ x \in [x_{i-1}, x_{i}],~i =1,2,\cdots,N.
	\end{align}
	The function, $f_{n}^{\alpha}$ is refereed as fractal interpolation function (FIF) associated with the iterated function system (IFS) $\left\{[a,b] \times \mathbf{C},\left(\pounds_{i}(x), \mathcal{J}_{i}(x, y)\right), i =1,2,\cdots,N\right\}$. We refer this function as the neural network $\alpha$-fractal function. This designation reflects the use of a neural network operator as the base function in the formulation of the IFS. Based on the methodology of construction of fractal functions as seen above, it can be demonstrated  that the Neural network $\alpha$-fractal function $f_{n}^{\alpha}$, of $f \in \mathcal{C}[a,b]$ for every $n \in \mathbb{N}$ is constructed using the IFS given by

	\begin{equation*}
		\Lambda_{i}=\left\{[a,b] \times \mathbf{C},\left(\pounds_{i}(x), \mathcal{J}_{i}(x, y)\right), i =1,2,\cdots,N\right\}, 
	\end{equation*}

	where $\pounds_i$ and $\mathcal{J}_{i}$ are defined by (\ref{defli}) and (\ref{deffi}) respectively. Since corresponding to any fixed partition $\Delta$ and scaling vector $\alpha$, we can get a unique  FIF $f^{\alpha}$ for each $f \in C[a,b]$, hence we can define an operator which takes each $f \in C[a,b]$ to corresponding $f^{\alpha}_{n}$ (or simply $f^{\alpha}$ ) as follows: \\
	\begin{definition}
		Let $f \in C[a,b]$, $\Delta$ be a fixed partition, $\alpha \in \mathbb{R}^N$ such that $|\alpha|_{\infty} <1$. Define the operator $\mathfrak{J}_{\Delta}^{\alpha}:C[a,b]\rightarrow C[a,b]$ by 
		\begin{align*}
			\mathfrak{J}_{\Delta}^{\alpha}(f)= f^{\alpha}_{n}
		\end{align*}
		where $f^{\alpha}_{n}$ is given by Equation (\ref{fif}).
		
	\end{definition}
	The operator defined above is termed as \textit{$\alpha$-fractal operator} with respect to partition $\Delta$ and scaling vector $\alpha$.
	From Equation (\ref{fif}), it is easy to notice that for a given $f \in \mathcal{C}[a,b]$ there is a sequence $\left\{f_{n}^{\alpha}(x)\right\}_{n=1}^{\infty}$ of Neural network $\alpha$-fractal functions.\\
	
	\begin{theorem}\label{Thm31}
		For any function $f\in\mathcal{C}[a,b]$, base function $\ss$ and partition $ \Delta$, the fractal function $f^\alpha$ converges uniformly to $f$ as $|\alpha|_\infty \to 0$, where $|\alpha|_\infty =\max \{\alpha_{i}, i=1,2,\cdots, N\}$.
		
	\end{theorem}
	
	\begin{proof}
		From (\ref{fif}), the uniform error arising from the $\alpha$-fractal approximation process can be estimated as follows:

		\begin{equation}\label{er}
			\left\|f^{\alpha}_{n}-f\right\|_{\infty} \leq \frac{|\alpha|_{\infty}}{1-|\alpha|_{\infty}}\|f-S_{n,\rho}(f, \cdot)\|_{\infty} .
		\end{equation}\\
		Thus from Equation (\ref{er}), for a given base function $\ss$ and partition \( \Delta \), \( f^\alpha \) converges uniformly to \( f \in \mathcal{C}[a, b] \) as \( |\alpha|_\infty \to 0 \).
	\end{proof}

	The next theorem discusses the convergence of the sequence \( \{f_n^\alpha(x)\}_{n=1}^\infty \) to \( f \in \mathcal{C}[a, b] \) for every possible choice of the scaling parameter \( \alpha \).\\

	\begin{theorem}\label{Thm3}
		For any function $f\in\mathcal{C}[a,b]$ and for every choice of scaling vector $\alpha=(\alpha_{1}, \alpha_{2}, \cdots, \alpha_{N})$, the sequence $\left\{f_{n}^{\alpha}(x)\right\}_{n=1}^{\infty}$ of neural network $\alpha$-fractal functions that converges uniformly to the original function $f$ as $n\rightarrow \infty$.
	\end{theorem}
	\begin{proof}
		From Equation (\ref{fif}), one can easily deduce 
		\begin{align*}
			\left\|f_{n}^{\alpha}-f\right\|_{\infty} & \leq|\alpha|_{\infty}\left\|f_{n}^{\alpha}-S_{n,\rho}(f, .)\right\|_{\infty} \\
			& \leq|\alpha|_{\infty}\left[\left\|f_{n}^{\alpha}-f\right\|_{\infty}+\left\|f-S_{n,\rho}(f, .)\right\|_{\infty}\right]\\
			&\leq \frac{|\alpha|_{\infty}}{1-|\alpha|_{\infty}}\left\|f-S_{n,\rho}(f, .)\right\|_{\infty},
		\end{align*}
		where $|\alpha|_\infty =\max \{\alpha_{i}, i=1,2,\cdots, N\}$.
		From Theorem 2 of \cite{QIAN2022126781}, we have 
		\begin{align*}
			\left\|f-S_{n,\rho}(f, .)\right\|_{\infty}\leq \omega(f,h),
		\end{align*}
		where $\omega(f,h)$ denotes the modulus of continuity\cite{Anastassiou2000} of $f$ with $h=\frac{b-a}{n}$.\\
		Hence, we get  
		\begin{align*}
			\left\|f_{n}^{\alpha}-f\right\|_{\infty} \leq \frac{|\alpha|_{\infty}}{1-|\alpha|_{\infty}}  \omega\left(f,h\right) 
		\end{align*}
		$f$ being  uniformly continuous on closed interval $[a,b]$, therefore $\omega\left(f,h\right)\rightarrow 0 $ as $h\rightarrow 0$.\\
	\end{proof}

	\begin{theorem}
		The $\alpha$-fractal operator $\mathfrak{J}_{n}^{\alpha}: \mathcal{C}[a,b] \rightarrow$ $\mathcal{C}[a,b]$ defined as $\mathfrak{J}_{n}^{\alpha}(f)=f_{n}^{\alpha}$ exhibits linearity and boundedness for every choice of scalar vector $\alpha$. Further $\mathfrak{J}_{n}^{\alpha}$ reduces to identity operator for $\alpha=0$.
	\end{theorem}
	
	\begin{proof}
		We have to prove that 
		\begin{align*}
			(\lambda f+\mu g)_{n}^{\alpha}=\lambda f_{n}^{\alpha}+\mu g_{n}^{\alpha}.
		\end{align*}
		As seen in the construction of $\alpha$-fractal function that $(af+bh)_{n}^{\alpha}$ is the fixed point of RB-operator $T^{\alpha}:X_{\beta_1}^{\beta_2}\rightarrow X_{\beta_1}^{\beta_2}$  given by
		\begin{align}\label{t}
			(T^\alpha h)(x)=\alpha_{i}h(\pounds_{i}^{-1}(x))+(\lambda f+\mu g)(x)-\alpha_{i}\mathcal{S}_{n, \rho}(\lambda f+\mu g,\pounds_{i}^{-1}(x))
		\end{align} 
		where $X_{\beta_1}^{\beta_2}=\left\{h\in C[a,b]:h(a)=\beta_1, h(b)=\beta_2\right\}$, $\beta_1=(\lambda f+\mu g)(a)$ and $\beta_2=(\lambda f+\mu g)(b)$.\\
		From Equation (\ref{fif}), we have $\forall~ i =1,2,\cdots,N$ and $\forall~  x \in [x_{i-1},x_i]$
		\begin{equation} \label{f}
			f_{n}^{\alpha}(x)=\alpha_{i} f_{n}^{\alpha}\left(\pounds_{i}^{-1}(x)\right)+f(x)-\alpha_{i} S_{n,\rho}\left(f, \pounds_{i}^{-1}(x)\right),
		\end{equation}
		\begin{equation} \label{g}
			g_{n}^{\alpha}(x)=\alpha_{i} g_{n}^{\alpha}\left(\pounds_{i}^{-1}(x)\right)+g(x)-\alpha_{i} S_{n,\rho}\left(g, \pounds_{i}^{-1}(x)\right).
		\end{equation}

		Multiplying Equation (\ref{f}) by $\lambda$ and Equation (\ref{g}) by $\mu$, then adding both the equations we have,
		\begin{align}\label{fg}
			(\lambda f_{n}^{\alpha}+\mu g_{n}^{\alpha})(x)=\alpha_{i}(f_{n}^{\alpha}+\mu g_{n}^{\alpha})(\pounds_{i}^{-1}(x))+(\lambda f+\mu g)(x)-\alpha_{i}(\mathcal{S}_{n,\rho}(\lambda f+\mu g), \pounds_{i}^{-1}(x))
		\end{align}
		Putting $h=\lambda f_{n}^{\alpha}+\mu g_{n}^{\alpha}$ in Equation (\ref{t}) and hence by uniqueness of fixed point and Equation (\ref{fg}), we have 
		\begin{equation*}
			(\lambda f+\mu g)_{n}^{\alpha}=\lambda f_{n}^{\alpha}+\mu g_{n}^{\alpha}, \quad \forall~ \lambda, \mu \in \mathbb{R},~~ \forall f,g \in C[a,b] ~ and ~~ \forall n \in \mathbb{N},
		\end{equation*}
		which proves the linearity of $\alpha$-fractal operator $\mathfrak{J}_{n}^{\alpha}$.\\
		For $\alpha=0$, Equation (\ref{f}) gives $f_{n}^{\alpha}=f$, i.e., $\mathfrak{J}_{n}^{0}=Id$, where $Id$ is identity operator.\\
		It remains to show that $\mathfrak{J}_{n}^{\alpha}$ is bounded. From Equation (\ref{er}), we have 
		\begin{align}\label{op}
			&\left\|f_{n}^{\alpha}\right\|_{\infty}-\|f\|_{\infty}\leq\frac{|\alpha|_{\infty}}{1-|\alpha|_{\infty}}\|f-S_{n,\rho}(f, \cdot)\|_{\infty} \nonumber\\
			\Rightarrow&\left\|f_{n}^{\alpha}\right\|_{\infty} \leq\|f\|_{\infty}+\frac{|\alpha|_{\infty}}{1-|\alpha|_{\infty}}\left\|Id-S_{n,\rho}(., .)\right\|_{\infty^{*}}\|f\|_{\infty} \nonumber\\
			or & \left\|\mathfrak{J}_{n}^{\alpha}(f)\right\|_{\infty} \leq\|f\|_{\infty}+\frac{|\alpha|_{\infty}}{1-|\alpha|_{\infty}}\left\|Id-S_{n,\rho}(., .)\right\|_{\infty^{*}}\|f\|_{\infty},
		\end{align}

		where \( \|\cdot\|_{\infty *} \) denotes the operator norm corresponding to the norm \( \|\cdot\|_{\infty} \).
		Also from \cite{QIAN2022126781}, the uniform convergence of sequence of operators $S_{n,\rho}(\cdot, \cdot)$ gives $\left\|Id-S_{n,\rho}(\cdot, \cdot)\right\|_{\infty^{*}} \rightarrow 0$ as $n \rightarrow \infty$. Thus, for given $\epsilon>0$, $\exists$ $\tilde{N} \in \mathbb{N}$ such that
		
		$$
		\left\|Id-S_{n,\rho}(\cdot, \cdot)\right\|_{\infty^{*}}<\epsilon, \quad \forall n>\tilde{N} .
		$$
		
		Let $M=\max \left\{\left\|Id-S_{1,\rho}(\cdot)\right\|_{\infty^{*}},\left\|Id-S_{2,\rho}(\cdot)\right\|_{\infty^{*}}, \ldots\left\|Id-S_{\tilde{N},\rho}(\cdot)\right\|_{\infty^{*}}, \epsilon\right\}$. Therefore, Equation (\ref{op}) gives
		\begin{align*}
			\left\|\mathfrak{J}_{n}^{\alpha}(f)\right\|_{\infty} \leq \left(1+\frac{|\alpha|_{\infty} M}{1-|\alpha|_{\infty}}\right)\|f\|_{\infty},
		\end{align*}
		which proves the boundedness of operator $\mathfrak{J}_{n}^{\alpha}$.
	\end{proof}
	\subsection{Numerical Experiments}
	To validate the theoretical results obtained, we provide some numerical examples with graphs as: 
	\begin{example}\label{ex1}
		Let the target function be $f(x)=2\sin(2x)$ and \newline $\Delta=[0, 0.11, 0.24, 0.35, 0.46, 0.57, 0.69, 0.79, 0.9, 1.0]$ be the partition of $[0,1]$. Let the sigmoidal function $\rho$ used in the NNOP $\mathcal{S}_{n,\rho}$ be
		
		\begin{align} \label{a1}
			\rho(x)=  \left\{
			\begin{array}{ll}
				0,\quad &x\leq -\frac{1}{2}  \\
				10(x+\frac{1}{2})^{3} -15(x+\frac{1}{2})^{4}+6(x+\frac{1}{2})^{5}, \quad & -\frac{1}{2} \leq x \leq \frac{1}{2} \\ 1, \quad& x \geq \frac{1}{2}
			\end{array}.
			\right.
		\end{align} 
		Figure \ref{f11} (a-d) represents the fractal functions $f_{n}^{n}$, with $n=7$ and $\alpha=0.8, 0.5, 0.3, 0.1$ respectively. It can be seen that the fractal function $f_{n}^{n}$ approaches to the original function with as $\alpha$ approaches to $0$ as the theoretical result obtained in Theorem \ref{Thm31}.

		\begin{figure}[htb!]
			\begin{subfigure}{.5\textwidth} 
				\includegraphics[width=1\linewidth, height=5cm]{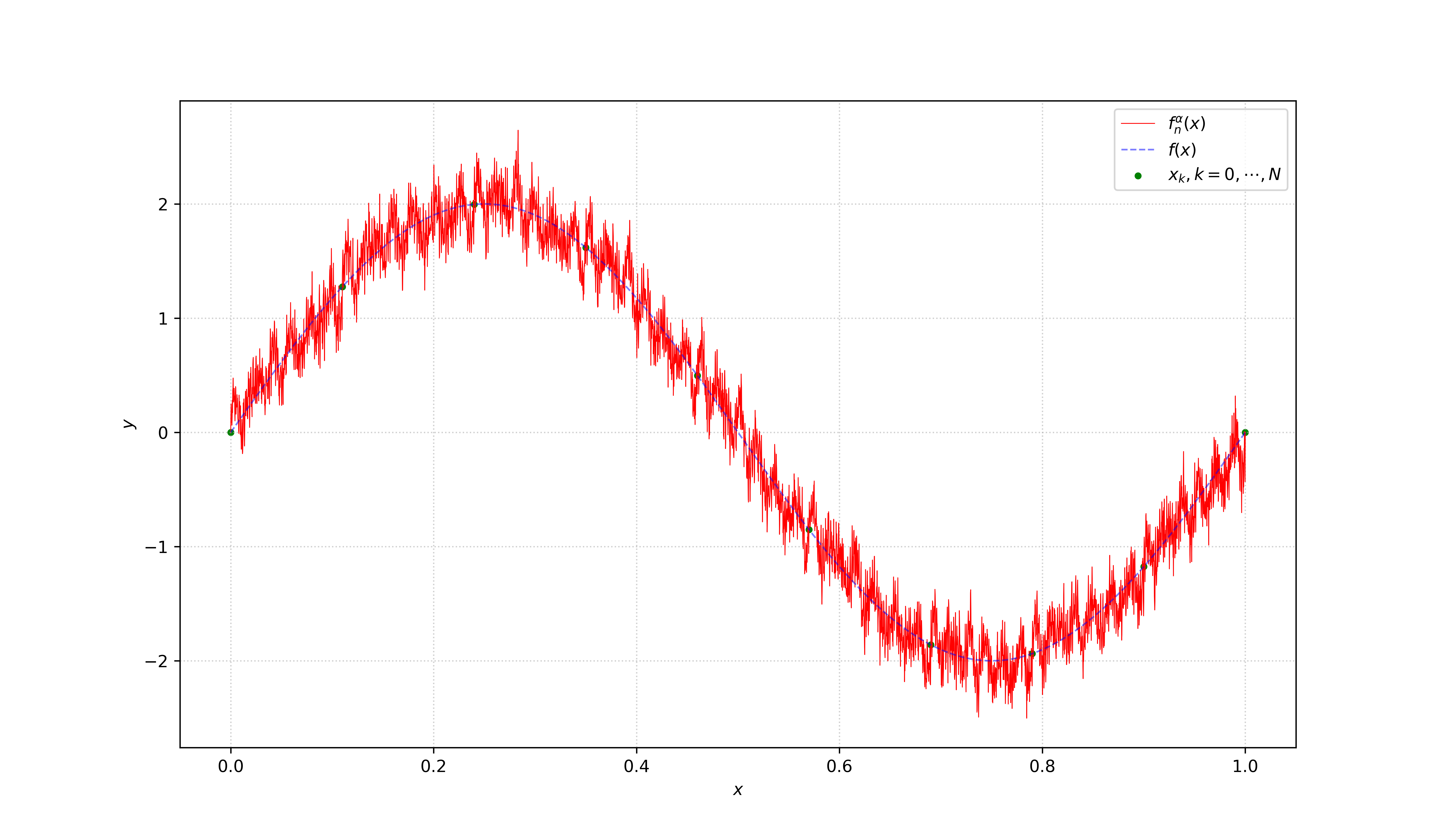} 
				\caption{$f_{7}^{0.8}$}
				\label{g1}
			\end{subfigure}
			\begin{subfigure}{.5\textwidth}
				\includegraphics[width=1\linewidth, height=5cm]{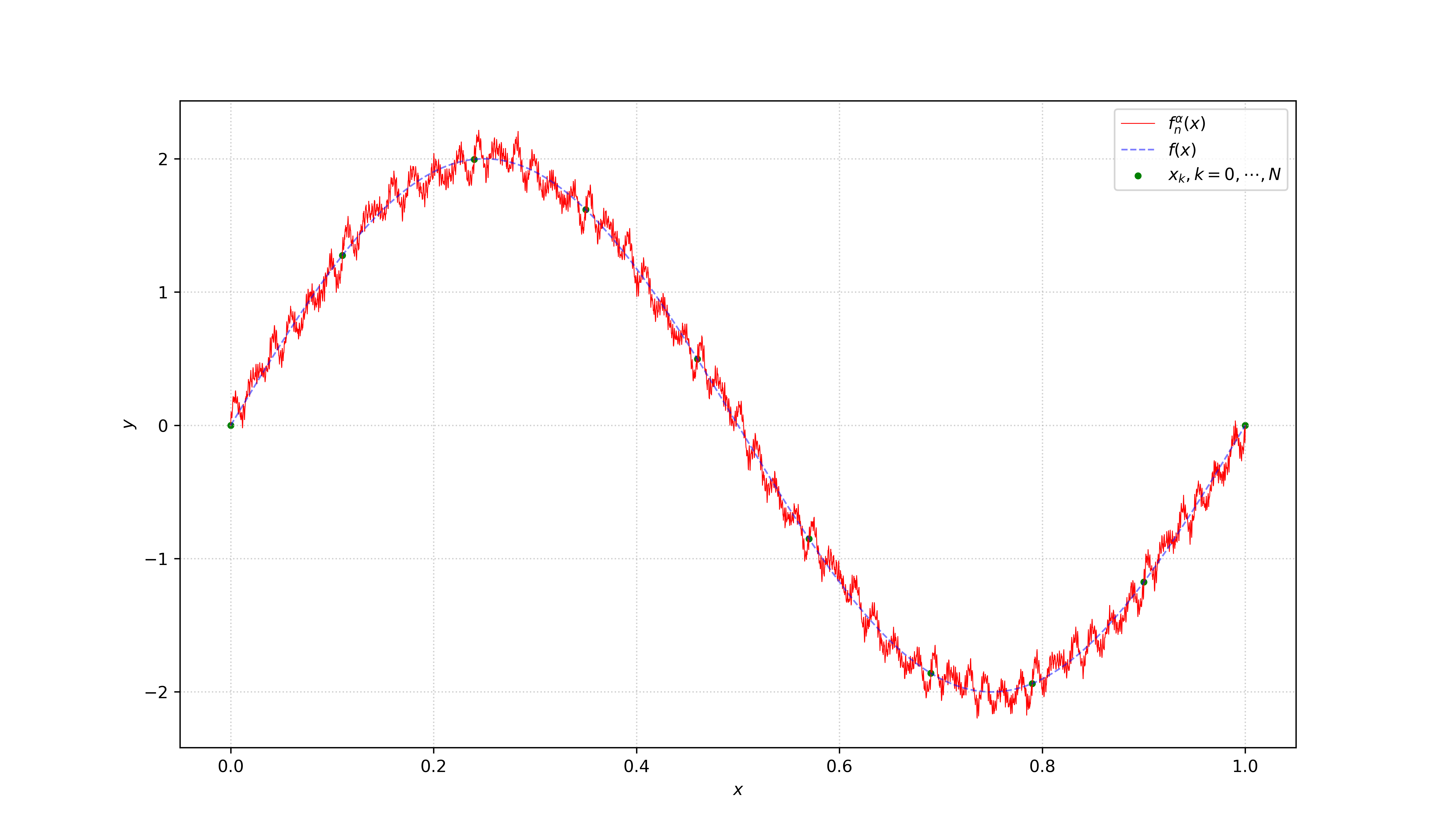}
				\caption{$f_{7}^{0.5}$}
				\label{g2}
			\end{subfigure}
			\begin{subfigure}{.5\textwidth}
				\includegraphics[width=1\linewidth, height=5cm]{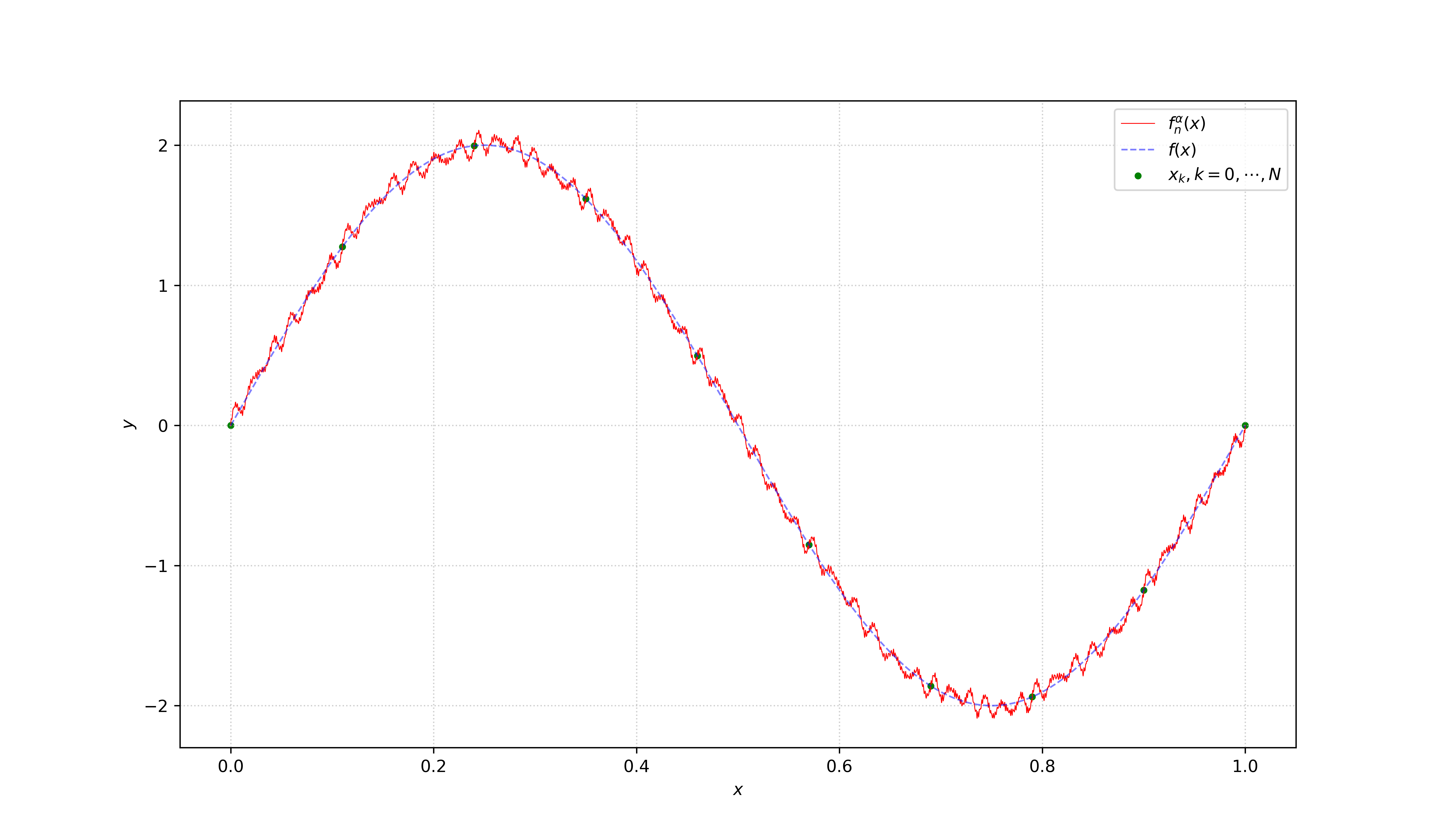}
				\caption{$f_{7}^{0.3}$}
				\label{g3}
			\end{subfigure}
			\begin{subfigure}{.5\textwidth}
				\includegraphics[width=1\linewidth, height=5cm]{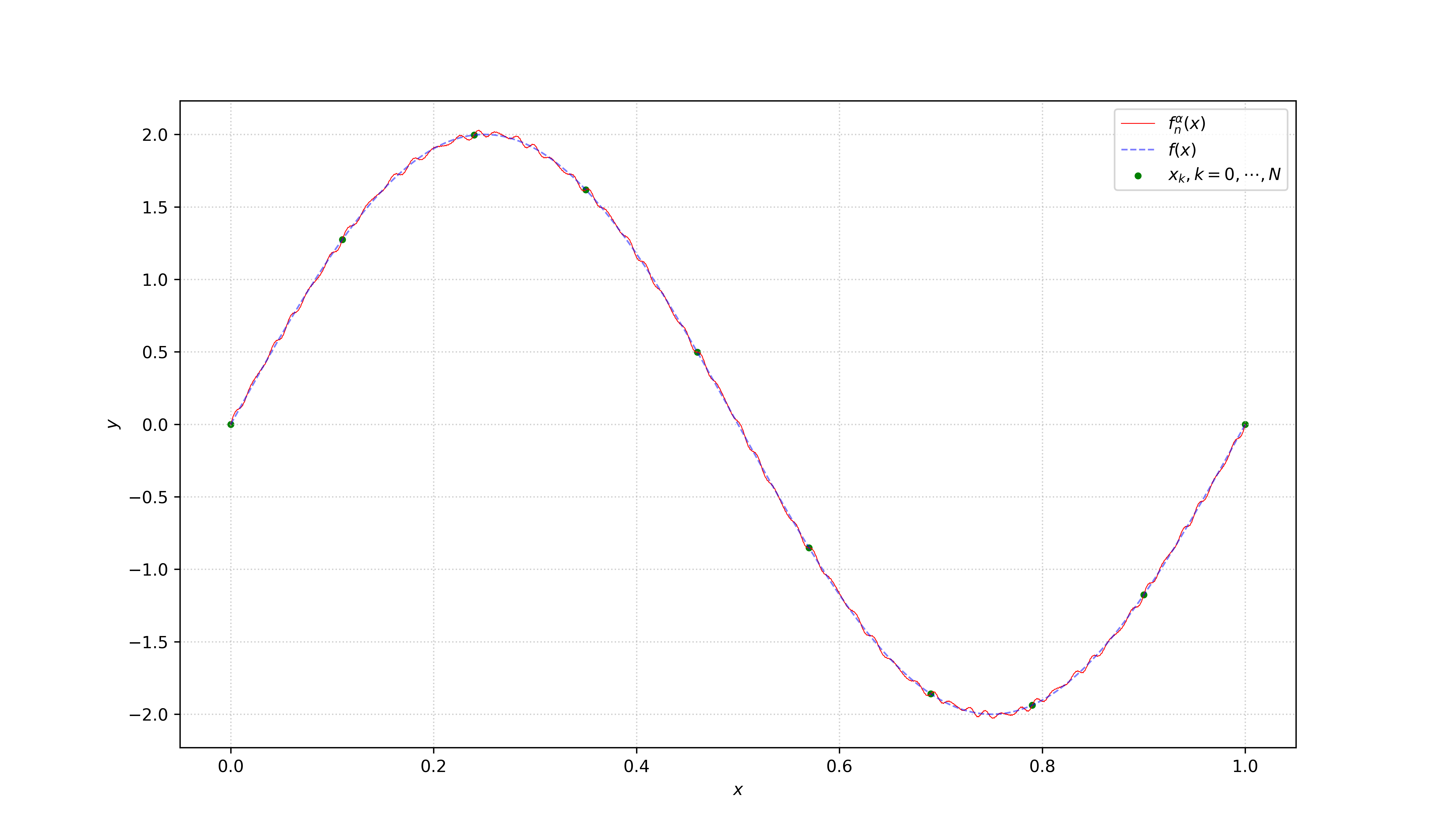}
				\caption{$f_{7}^{0.1}$}
				\label{g4}
			\end{subfigure}
			
			\caption{Comparison among fractal interpolation functions for fixed $n$ and varying $\alpha$}\label{f11}
		\end{figure}

		Figure \ref{f12} (a-d) represents the fractal functions $f_{n}^{\alpha}$, with $\alpha=0.4$ and $n=3, 5, 9, 15$ respectively. It can be seen that the fractal function $f_{n}^{\alpha}$ approaches to the original function with increase in the value of $n$ as the theoretical result obtained in Theorem \ref{Thm3}.

		\begin{figure}[htb!]
			\begin{subfigure}{.5\textwidth} 
				\includegraphics[width=1\linewidth, height=5cm]{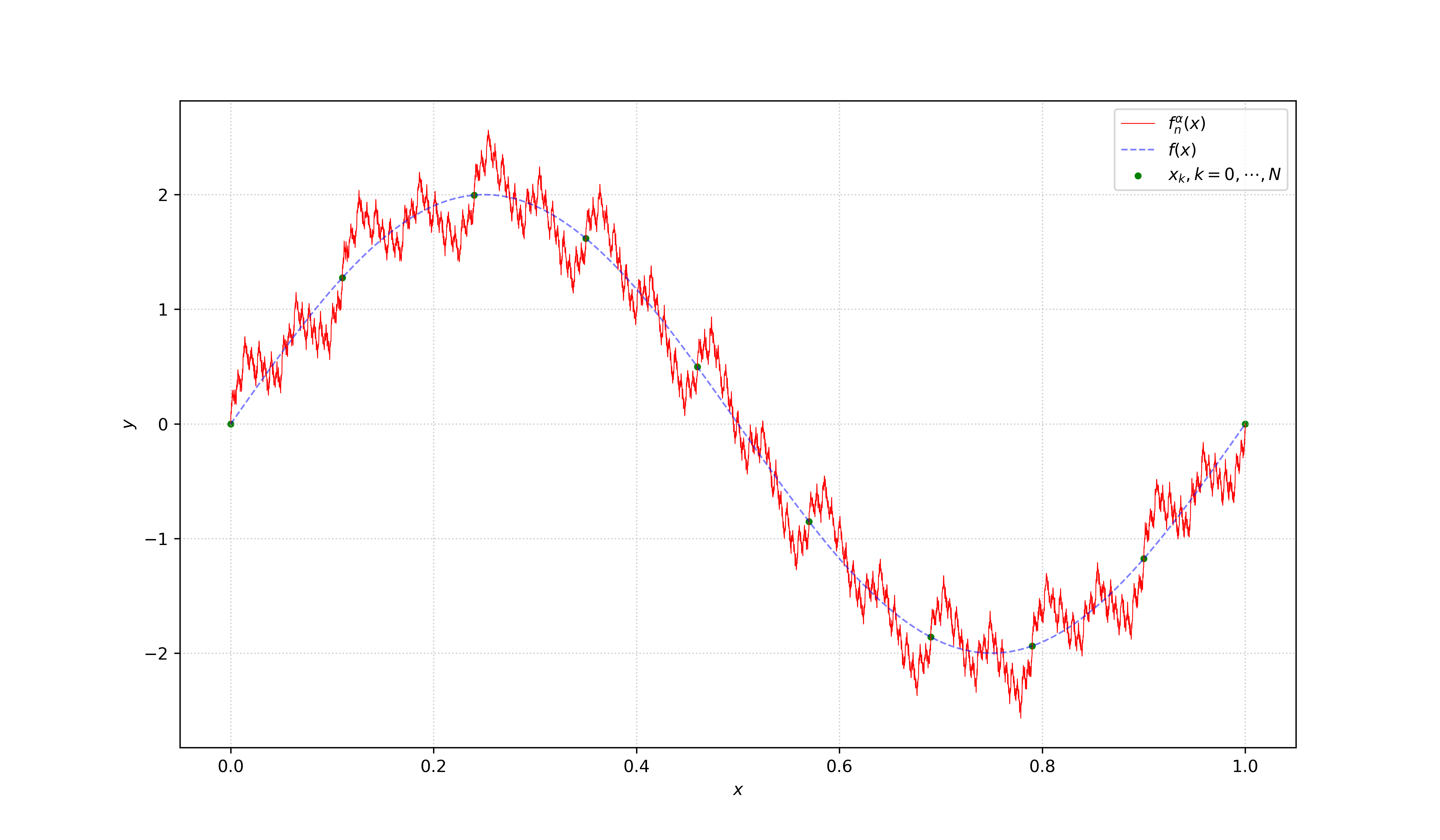} 
				\caption{$f_{3}^{0.4}$}
				\label{g12}
			\end{subfigure}
			\begin{subfigure}{.5\textwidth}
				\includegraphics[width=1\linewidth, height=5cm]{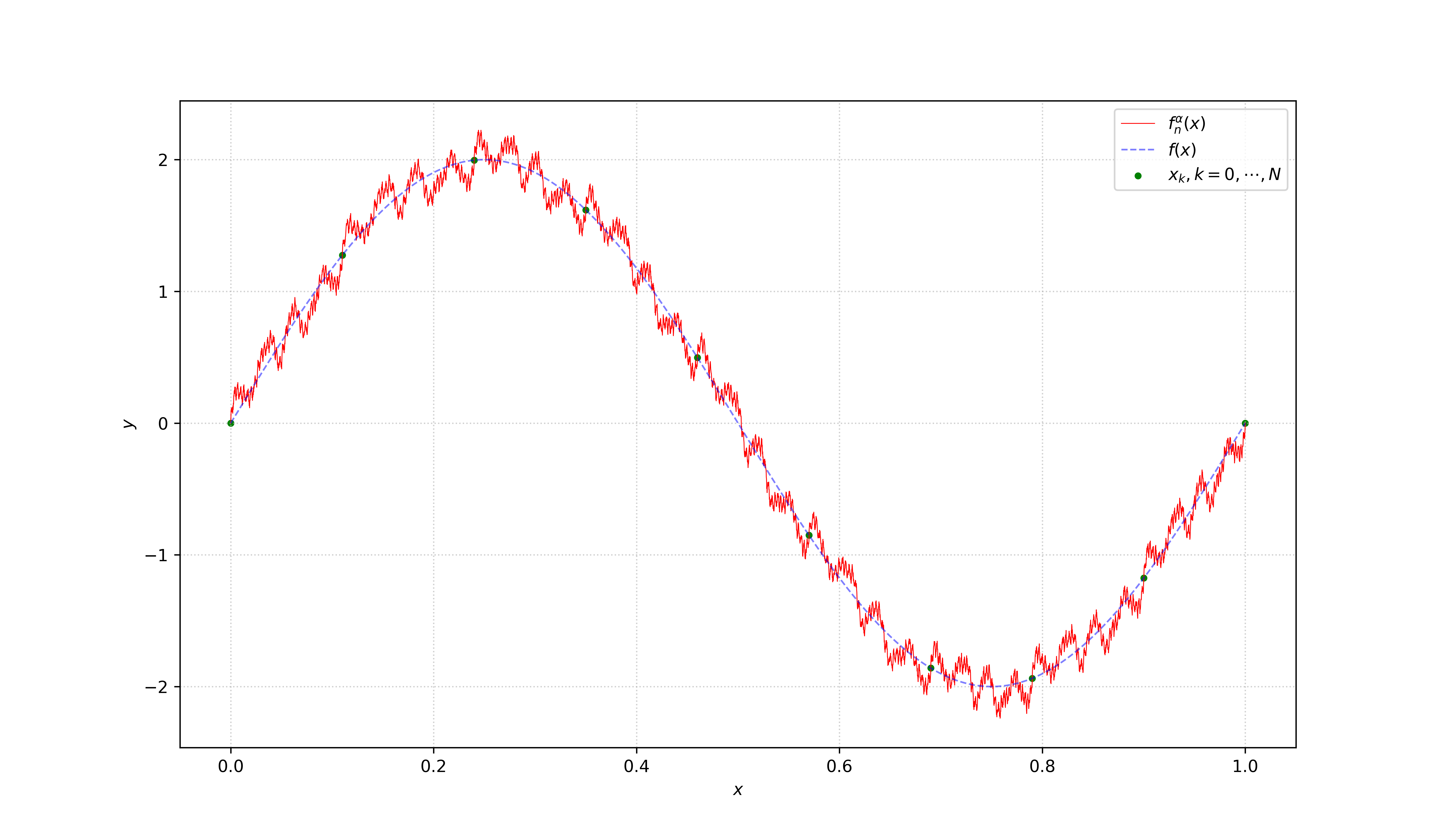}
				\caption{$f_{5}^{0.4}$}
				\label{g22}
			\end{subfigure}
			\begin{subfigure}{.5\textwidth}
				\includegraphics[width=1\linewidth, height=5cm]{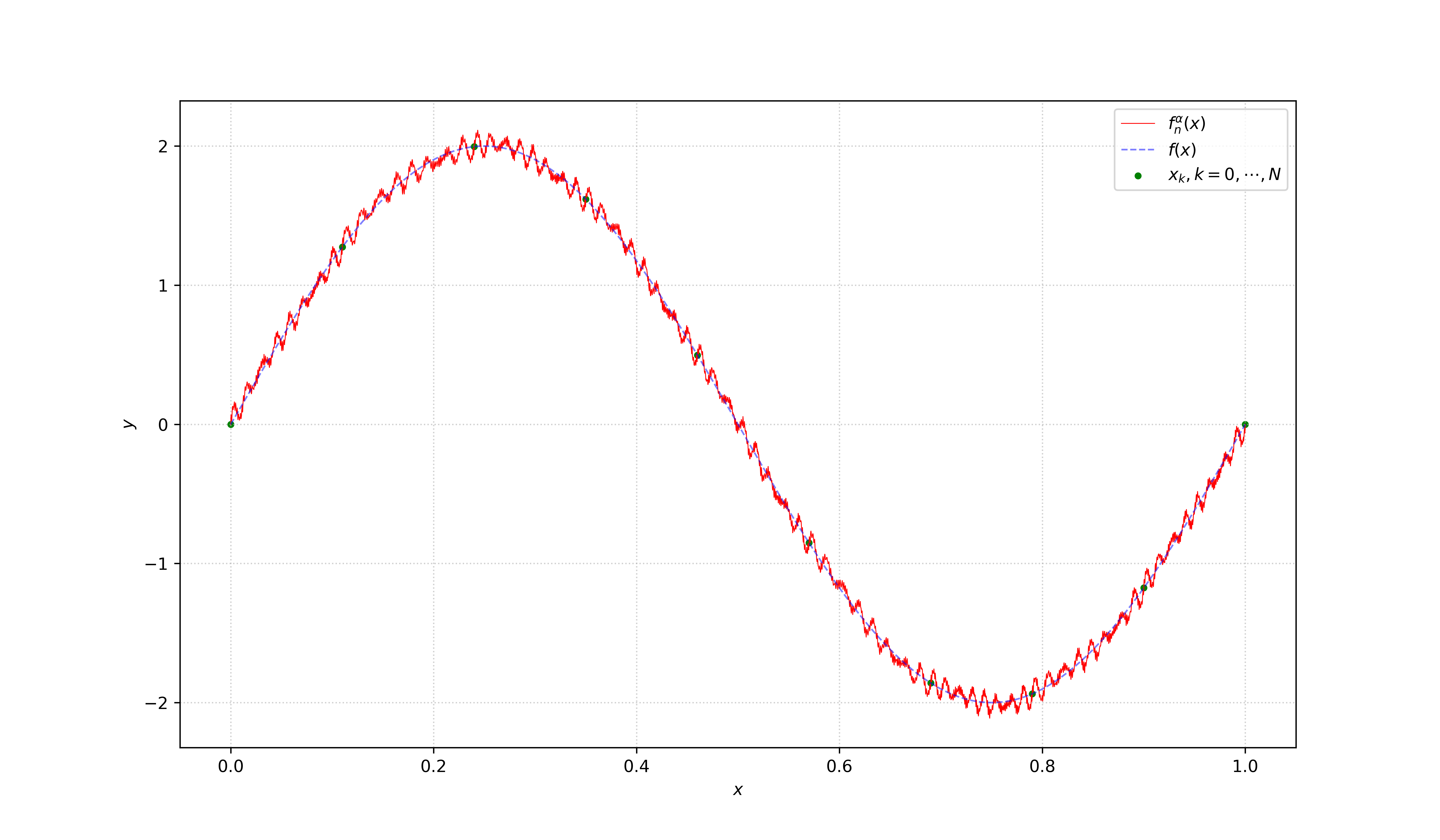}
				\caption{$f_{9}^{0.4}$}
				\label{g32}
			\end{subfigure}
			\begin{subfigure}{.5\textwidth}
				\includegraphics[width=1\linewidth, height=5cm]{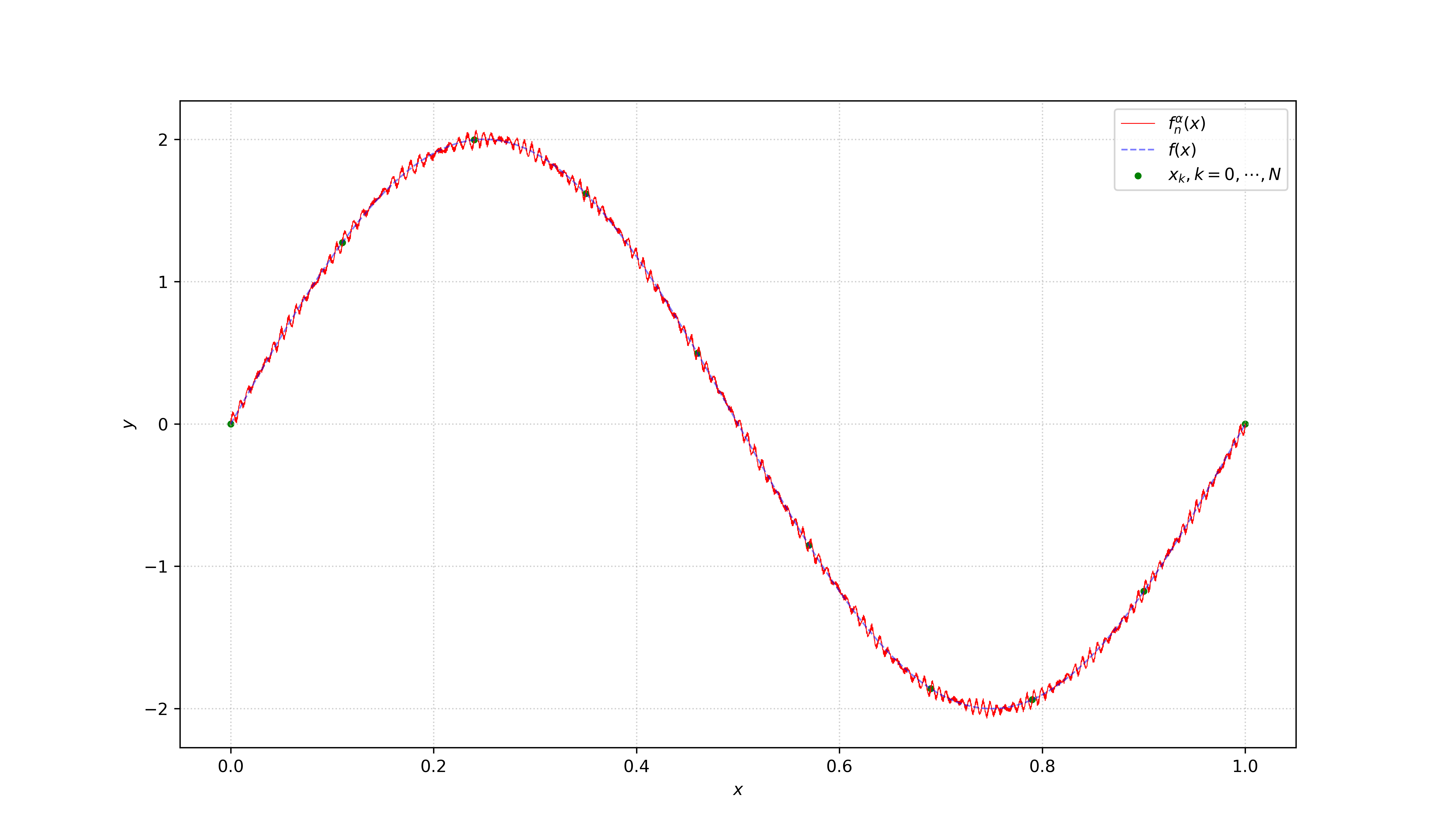}
				\caption{$f_{15}^{0.4}$}
				\label{g42}
			\end{subfigure}
			
			\caption{Comparison among fractal interpolation functions for fixed $\alpha$ and varying $n$}\label{f12}
		\end{figure}
	\end{example}

	\begin{example}[Approximating the Weierstrass Function]
		In this example, we compare the approximation of Weierstrass function by ordinary neural network operators and neural network based fractal interpolations functions. Let the partition $\Delta$ and sigmoidal function $\rho$ be same as in Example \ref{ex1}. The Weierstrass Function is defined as: \begin{align*}
			g(x) = \sum_{\eta=0}^{\infty} \gamma^\eta \cos(\beta^\eta \pi x),\\
		\end{align*}
		where, $0 < \gamma < 1$, $\beta$ is a positive odd integer, and
		$\gamma \beta > 1 + \frac{3}{2}\pi$.
		
		Take $\gamma=0.5$, $\beta=13$.
		
		\begin{figure}[htb!]
			\begin{subfigure}{.5\textwidth} 
				\includegraphics[width=1\linewidth, height=5cm]{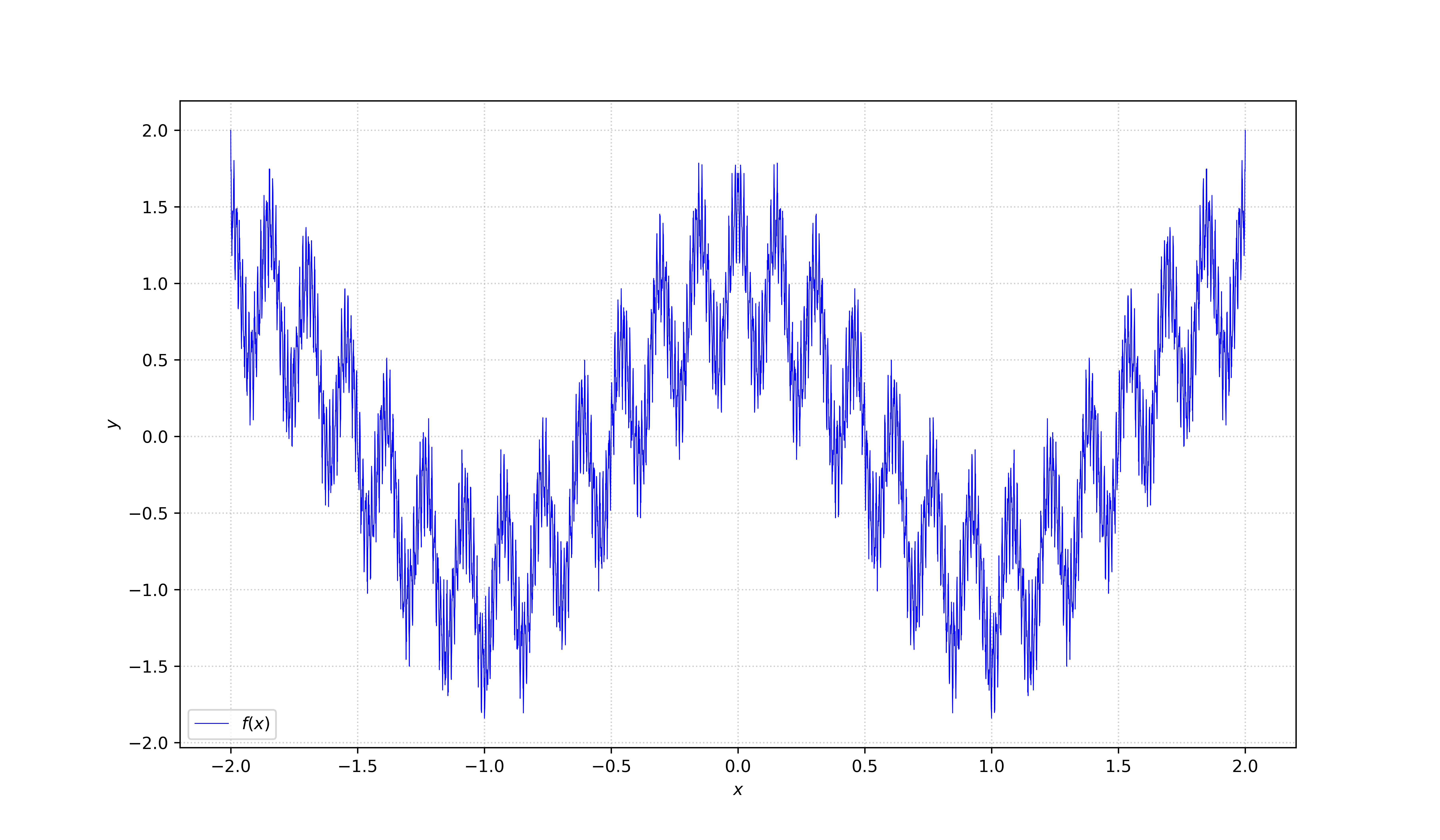} 
				\caption{$Weierstrass function (g)$}
				\label{w1}
			\end{subfigure}
			\begin{subfigure}{.5\textwidth}
				\includegraphics[width=1\linewidth, height=5cm]{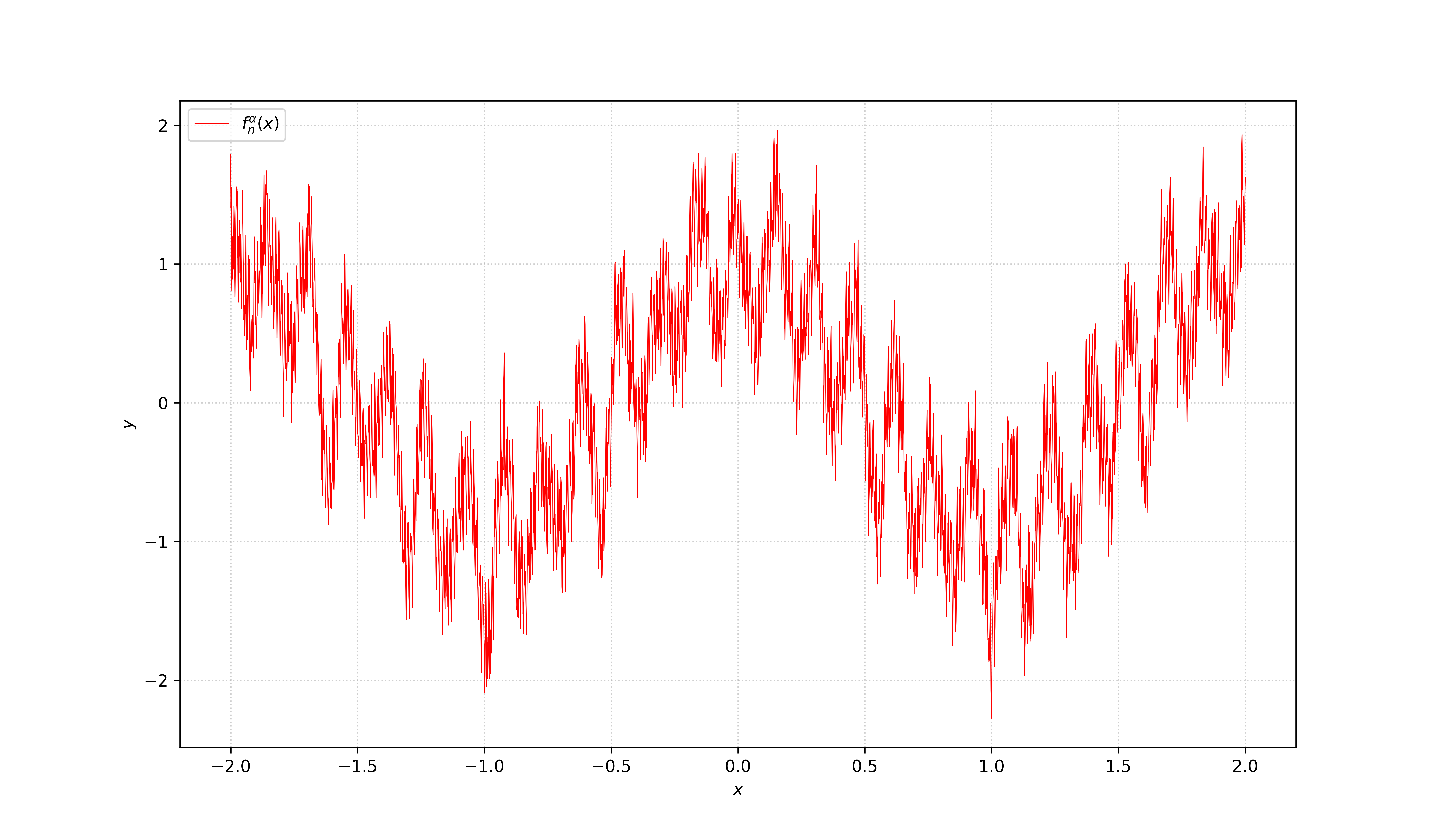}
				\caption{$g_{10}^{0.3}$}
				\label{w2}
			\end{subfigure}
			\begin{subfigure}{.5\textwidth}
				\includegraphics[width=1\linewidth, height=5cm]{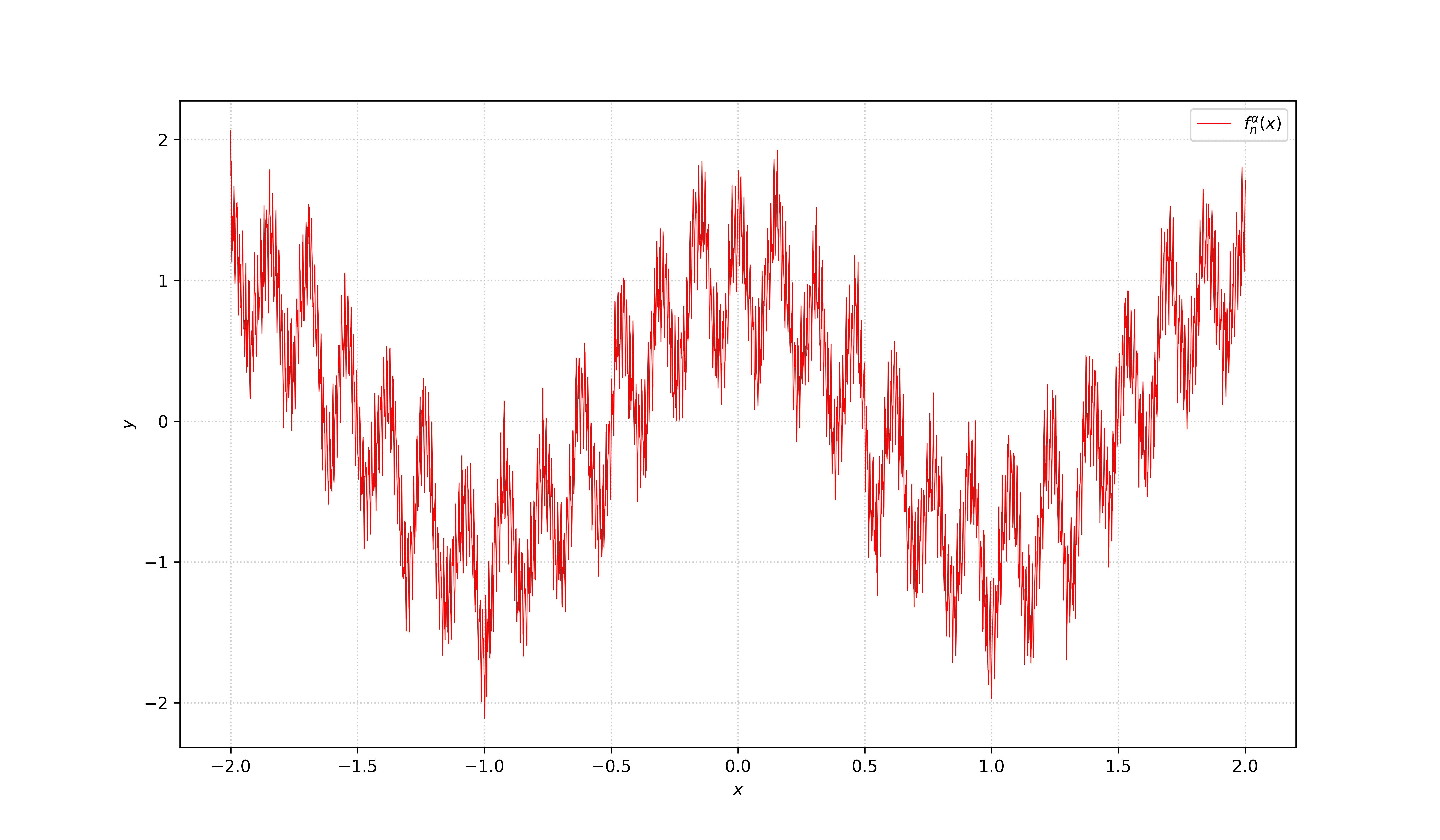}
				\caption{$g_{15}^{0.2}$}
				\label{w3}
			\end{subfigure}
			\begin{subfigure}{.5\textwidth}
				\includegraphics[width=1\linewidth, height=5cm]{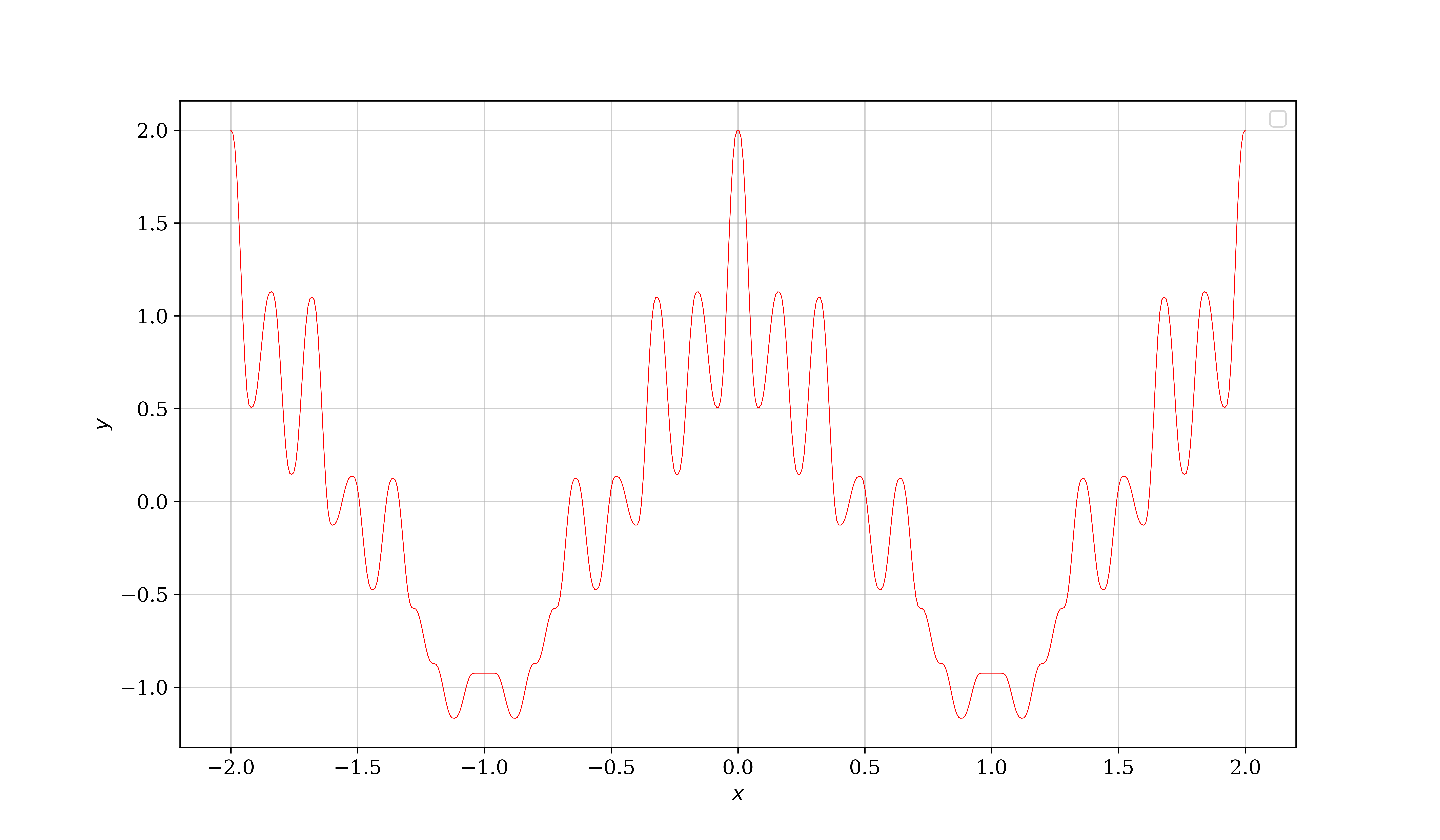}
				\caption{$\mathcal{S}_{50,\rho}(g)$}
				\label{w4}
			\end{subfigure}
			
			\caption{Comparison between fractal approximation and neural network approximation}\label{w}
		\end{figure}
		
		Figure \ref{w1}, \ref{w2} and \ref{w3} represents the Weierstrass function $g$, the approximation by neural network operator $\mathcal{S}_{\rho, n}(g)$ and fractal interpolation function $g_{n}^{\alpha}$ respectively. It is clear from figure \ref{f12} that the fractal interpolation function approximates the Weierstrass function $g$ more accurately than an ordinary neural network operator, emphasizing the importance of approximation by fractal interpolation functions.
	\end{example}
	
	\section{Neural Network $\alpha$-fractal function in H\"{o}lderian spaces.}\label{sec4}
	
	The fractal dimension of \(f^{\boldsymbol{\alpha}}\) is determined by the scaling vector \(\boldsymbol{\alpha}\).
	Nasim et al. \cite{nasim} calculated the box dimension of the \(\alpha\)-fractal functions. Their work on the box dimension of $\alpha$-FIFs relies on carefully controlling the properties of the original function, the base function, and the scaling function. By imposing the appropriate conditions on these functions, they calculated the box dimension of $\alpha$-fractal interpolation function. The next proposition presents the detailed formulation of their approach.\\
	\begin{proposition}\label{naseem}
		
		Let $f \in \mathcal{C}[a,b]$ and base function $\ss: [a,b] \rightarrow \mathbb{R}$ be H$\ddot{o}$lder continuous functions satisfying the condition (\ref{b1}). For the partition  $\Delta=\left\{x_{0}, x_{1}, x_{2}, \ldots, x_{N}\right\}$ of [a,b] satisfying $x_{0}<x_{1}<\cdots<x_{N}$ and $\alpha=\left(\alpha_{1}, \alpha_{2}, \ldots, \alpha_{N}\right)$, with $|\alpha_{i}|<1, \forall~i=1,2,\cdots,N$. If the data points $\left(x_{i}, f\left(x_{i}\right)\right), i =0, 1,2,\cdots,N$ are noncollinear, then the box dimension $D$ of the graph corresponding to the  $\alpha$-fractal function $f^{\alpha}$ satisfies the following:
		
		\begin{align*}
			D= \begin{cases}1+\log _{N} \kappa, & \text { if } \kappa>1 \\ 1, & \text { otherwise }\end{cases}
		\end{align*}
		
		where $\kappa=\displaystyle \sum_{i=1}^{N}\left|\alpha_{i}\right|$.
	\end{proposition} 
	
	\begin{definition}
		For $0<\mu \leq 1$, let

		\begin{equation*}
			\mathcal{H}^{0,\mu}=\left\{\phi: [a,b] \rightarrow \mathbb{R}: \sup _{x \neq y} \frac{|\phi(x)-\phi(y)|}{|x-y|^{\mu}}<\infty\right\}.
		\end{equation*}
		
		Then, $\mathcal{H}^{0,\mu}$ forms a vector space called H\"{o}lderian Space or Lipschitz Space.\\
		Define $\|\phi\|_{0, \mu}=\max \left\{\|\phi\|_{\infty},|\phi|_{\mu}\right\}$, where
		\begin{equation*}
			|\phi|_{\mu}=\sup \left\{\frac{|\phi(x)-\phi(y)|}{|x-y|^{\mu}}, x, y \in [a,b], x \neq y\right\}.
		\end{equation*}
	\end{definition}
	$\left(\mathcal{H}^{0,\mu},\|\cdot\|_{0, \mu}\right)$ forms a complete metric space.\\
	
	The assumptions in Proposition \ref{naseem} require that both the functions $f$ and 
	$\ss$ to be H\"{o}lderian. However, the base function $\ss=S_{n,\rho}f$,  does not necessarily inherit H\"{o}lder continuity from $f$.  To ensure that the base function is H\"{o}lder continuous, we impose the condition that the sigmoidal function $\rho$ (\ref{sig}), used in the activation function while defining the neural network operator $S_{n,\rho}f$ \cite{QIAN2022126781}, is H\"{o}lder continuous, which ensures that $S_{n,\rho}f$ H\"{o}lderian as well. In the next theorem we will develop an $\boldsymbol{\alpha}$-fractal function corresponding to the given function $f\in \mathcal{H}^{0,\mu}$.
	
	\begin{theorem}\label{Thml}
		Let $f \in \mathcal{H}^{0,\mu}$. Suppose $\Delta=\left\{a=x_{0}, x_{1}, x_{2}, \ldots, x_{N}=b\right\}$ is a partition of closed interval $[a,b]$ with $x_{0}<x_{1}<\cdots<x_{N}$. For all $i =1,2,\cdots,N$, let $\pounds_{i}$ be given by (\ref{defli}) and $\mathcal{J}_i$ be defined by (\ref{deffi}) with $\rho \in \mathcal{H}^{0,\mu}$ and scaling function
		$\alpha_{i} \in \mathcal{H}^{0,\mu}$. Let the RB-operator $T:\left(\mathcal{H}^{0,\mu},\|\cdot\|_{0, \mu}\right) \rightarrow\left(\mathcal{H}^{0,\mu},\|\cdot\|_{0, \mu}\right)$  be defined by:
		
		\begin{equation*}
			T\phi(x)=f(x)+\boldsymbol{\alpha}_{i}\left(\pounds_{i}^{-1}(x)\right)\left(\phi-S_{n,\rho} f\right)\left(\pounds_{i}^{-1}(x)\right).
		\end{equation*}
		
		\noindent If the scaling vector $\alpha$ satisfies 
		\begin{equation}\label{scale}
			\max \left\{\frac{\left\|\alpha_{i}\right\|_{\infty}}{a_{i}^{\mu}}: i =1,2,\cdots,N\right\}<1,
		\end{equation} then  $T$ admits a unique fixed point $f^{\alpha} \in \mathcal{H}^{0,\mu}$.\\
	\end{theorem}
	
	\begin{proof}
		
		Consider an arbitrary element $\phi \in \mathcal{H}^{0,\mu}$. Our aim is to show that $T\phi \in \mathcal{H}^{0,\mu} $. For each $n \in \mathbb{N}$,
		\begin{align*}
			|T \phi|_{\mu}= & \sup _{\underset{x \neq y}{x, y \in [a,b],}} \frac{|T \phi(x)-T \phi(y)|}{|x-y|^{\mu}} \\
			= & \sup _{\underset{x \neq y}{x, y \in [x_{i-1},x_i], }} \frac{\left|f(x)-f(y)+\alpha_{i}\left(\phi-S_{n,\rho}(f)\right) \circ\left(\pounds_{i}^{-1}(x)\right)-\alpha_{i}\left(\phi-S_{n,\rho}(f)\right) \circ\left(\pounds_{i}^{-1}(y)\right)\right|}{|x-y|^{\mu}} \\
			\leq & \sup _{\underset{x \neq y}{x, y \in [x_{i-1},x_i], }}\left\{\frac{|f(x)-f(y)|}{|x-y|^{\mu}}\right\}+\max _{i =1,2,\cdots,N}\left(\left\|\alpha_{i}\right\|_{\infty}\right) \sup _{\underset{x \neq y}{x, y \in [x_{i-1},x_i], }}\left[\frac{\left|\phi\left(\pounds_{i}^{-1}(x)\right)-\phi\left(\pounds_{i}^{-1}(y)\right)\right|}{|x-y|^{\mu}}\right. \\
			& \left.+\frac{\left|S_{n,\rho}\left(f ; \pounds_{i}^{-1}(x)\right)-S_{n,\rho}\left(f ; \pounds_{i}^{-1}(y)\right)\right|}{|x-y|^{\mu}}\right]
		\end{align*}

		From assumptions $f \in \mathcal{H}^{0,\mu},~ S_{n,\rho}(f) \in \mathcal{H}^{0,\mu}$, the above inequality implies $|T \phi|_{\mu}<\infty$, and hence $T \phi \in \mathcal{H}^{0,\mu}$.\\
		Our next claim is to prove that $T$ is contraction. For this, let $\phi,~ \psi \in \mathcal{H}^{0,\mu}$, we have
		
		\begin{equation}\label{10}
			\left\|T\phi - T\psi\right\|_{\infty} \leq \max _{i =1,2,\cdots,N}\left(\left\|\alpha_{i}\right\|_{\infty}\right)\left\|T\phi - T\psi\right\|_{\infty}
		\end{equation}

		Following a similar approach as in the estimation of $|T g|_{\mu}$, we derive
		
		\begin{equation}\label{11}
			\left|T\phi - T\psi\right|_{\mu} \leq \max _{i =1,2,\cdots,N}\left(\frac{\left\|\alpha_{i}\right\|_{\infty}}{a_{i}^{\mu}}\right)\left|\phi - \psi\right|_{\mu}
		\end{equation}
		
		From the inequalities (\ref{10}) and (\ref{11}), we get
		\begin{align}\label{contr}
			\left\|T\phi - T\psi\right\|_{0, \mu} & =\max \left\{\left\|T\phi - T\psi\right\|_{\infty},\left|T\phi - T\psi\right|_{\mu}\right\}\nonumber \\
			& \leq \max \left\{\max _{i =1,2,\cdots,N}\left(\left\|\alpha_{i}\right\|_{\infty}\right)\left\|\phi - \psi\right\|_{\infty}, \max _{i =1,2,\cdots,N}\left(\frac{\left\|\alpha_{i}\right\|_{\infty}}{a_{i}^{\mu}}\right)\left|\phi - \psi\right|_{\mu}\right\} \nonumber \\
			& =\max _{i =1,2,\cdots,N}\left(\frac{\left\|\alpha_{i}\right\|_{\infty}}{a_{i}^{\mu}}\right) \max \left\{\left\|\phi - \psi\right\|_{\infty},\left|\phi - \psi\right|_{\mu}\right\} \nonumber\\
			& =\max _{i =1,2,\cdots,N}\left(\frac{\left\|\alpha_{i}\right\|_{\infty}}{a_{i}^{\mu}}\right)\left\|\phi - \psi\right\|_{0, \mu} .
		\end{align}
		
		Hence, from assumption (\ref{scale}) and from inequality given in (\ref{contr}), the operator $T$ is a contraction defined on a complete metric space $\left(\mathcal{H}^{0,\mu}, \|\cdot\|_{0, \mu}\right)$. Therefore, by Banach contraction principle, $T$ admits a unique fixed point, say $f_{n}^{\alpha}$. \\
	\end{proof}
	
	\begin{theorem}	
		Let $f \in \mathcal{H}^{0,\mu}$. Under the assumptions of the Theorem \ref{Thml} the sequence of $\boldsymbol{\alpha}$-fractal functions $\left\{f_{n}^{\alpha}\right\}_{n=1}^{\infty}$ converges to the target function $f$ as $n \rightarrow \infty$.\\
	\end{theorem}
	
	\begin{proof}
		
		By applying a similar procedure as in Theorem \ref{Thm3}, we can conclude
		\begin{align}\label{100}
			\left\|f_{n}^{\alpha}-f\right\|_{0, \mu} \leq \frac{M}{1-M}\left\|f-S_{n,\rho}(f)\right\|_{0, \mu},
		\end{align}
		where  $M=\max \left\{\frac{\left\|\alpha_{i}\right\|_{\infty}}{a_{i}^{\mu}}: i =1,2,\cdots,N\right\}<1 $
		
		As established in \cite{QIAN2022126781}, we have
		\begin{equation}\label{110}
			\left\|f-S_{n,\rho}(f)\right\|_{0, \mu} \rightarrow 0 \text { as } n \rightarrow \infty 
		\end{equation}
		
		Therefore, the proof is established directly from Equations (\ref{100}) and (\ref{110}).\\
	\end{proof}

	\section{Fractal interpolation functions using a discrete approximation approach} \label{sec5}
	We have observed in the Equation (\ref{fif}) that the $\alpha$-fractal $f_{n}^{\alpha}$ depends on all values of the target function $f$, which is not a good choice for approximation or interpolation, particularly when considering computational efficiency and other practical factors. To overcome these limitations  a discrete approximation framework is introduced for fractal approximation. In order to reduce the computational cost, while maintaining the essential features of the target function, our method computes the FIF on a finite set of discrete values. We consider the uniform spaced partition $\Delta= \{x_0, x_1, \cdots, x_N\}$, where $x_k = a+k\tilde{h}, ~ k=0,1,\cdots, N$ and $\tilde{h}=\frac{b-a}{N}$.
	From Equation (\ref{qi}) of Section \ref{sec2}, it is evident that there is flexibility in choosing the height function $h$. The only criterion is that it satisfies the property (\ref{hi}). In this section we consider $h$ as following:
	
	\begin{equation}
		h(x)=S_{N,\rho}(f;x)
	\end{equation}
	where 
	\begin{align}
		S_{N,\rho}(f, x)= \sum_{k=0}^{N}f(x_k)\xi\left(\frac{2m}{\tilde{h}}(x-x_k)\right),\quad \forall~ x \in [a,b], \quad\forall ~N \in \mathbb{N} . 
	\end{align}
	where $\tilde{h}=\frac{b-a}{N}$, and $x_k=a+k\tilde{h}$, $k=1,2,\cdots, N$.\\
	
	Now, we define the IFS through the following functions:
	
	\begin{equation}\label{defli2}
		\pounds_i(x)= \frac{x_i -x_{i-1}}{x_N-x_0}.x +\frac{x_N x_{i-1} -x_0 x_i}{x_N-x_0}, \quad i =1,2,\cdots,N
	\end{equation}\\
	and 
	\begin{equation}\label{deffi2}
		\mathcal{J}_i(x,y)=\alpha_{i}.y+S_{N,\rho}(f,\pounds(x))-\alpha_{i}S_{n,\rho}(f, x), \quad i =1,2,\cdots,N.
	\end{equation}\\
	$\pounds_i$ satisfy the conditions (\ref{ln}) as discussed earlier. It remains to show that $\mathcal{J}_i$ satisfy the condition (\ref{fn}).\\
	Using the interpolation properties of $S_{N,\rho}(f,(x))$ \cite{QIAN2022126781}, we have for $ \forall i=1,2,\cdots,N$
	\begin{align*}
		\mathcal{J}_i(x_0,f(x_0))&=\alpha_{i}.f(x_0)+S_{N,\rho}(f,\pounds_{i}(x_0))-\alpha_{i}S_{n,\rho}(f, x_0)\\
		&=\alpha_{i}.f(x_0)+S_{N,\rho}(f,x_{i-1})-\alpha_{i}.f(x_0)\\
		&=f(x_{i-1})
	\end{align*}\\
	Similarly, one can show that $\forall~ i=1,2,\cdots,N$,
	\begin{align*}
		\mathcal{J}_i(x_N,f(x_N))=f(x_i).
	\end{align*}\\
	Also $\mathcal{J}_i$ satisfies the Lipschitz condition in second argument i.e.,
	\begin{align*}
		|\mathcal{J}_i(x,y)-\mathcal{J}_i(x,z)|\leq |\alpha_{i}||y-z|.
	\end{align*}
	By following the similar steps as we did in the construction of $\alpha$-fractal approximation in Section \ref{sec3}, we easily derive Fractal Interpolation Function (FIF), say $f_{n, N}^{\alpha}$, given by
	\begin{align}\label{fif2}
		f_{n, N}^{\alpha}(x)=\alpha_{i}f_{n, N}^{\alpha}(\pounds_{i}^{-1}(x))+S_{N,\rho}(f, x)-\alpha_{i}S_{n,\rho}(f, \pounds_{i}^{-1}(x)),~ x\in [x_{i-1}, x_i],
	\end{align}
	$  \forall i=1,2,\cdots,N.$
	It is evident from Equation (\ref{fif2}) that the fractal interpolation function $f_{n, N}^{\alpha}$ depends solely on the function values at the partition points (nodes), unlike methods developed previously that rely on each value attained by the target function.\\  
	Also, from (\ref{fif2}), the uniform error bound can be easily calculated as follows:
	\begin{align*}
		|f_{n, N}^{\alpha}(x)-f(x)|&=|\alpha_{i}f_{n}^{\alpha}(\pounds_{i}^{-1}(x))+S_{N,\rho}(f, x)-\alpha_{i}S_{n,\rho}(f, \pounds_{i}^{-1}(x))-f(x)|\\
		&\leq |\alpha|_\infty \|f_{\alpha}^{n}- S_{n,\rho}(f,\cdot)\|_{\infty}+\|S_{N,\rho}(f,\cdot)-f\|_{\infty}\\
		&\leq |\alpha|_\infty (\|f_{\alpha}^{n}-f\|_{\infty}+\|f- S_{n,\rho}(f,\cdot)\|_{\infty})+\|S_{N,\rho}(f,\cdot)-f\|_{\infty}
	\end{align*}
	which implies
	\begin{align}\label{omega}
		\|f_{n, N}^{\alpha}-f\|_{\infty}&\leq|\alpha|_\infty (\|f_{\alpha}^{n}-f\|_{\infty}+\|f- S_{n,\rho}(f,\cdot)\|_{\infty})+\|S_{N,\rho}(f,\cdot)-f\|_{\infty} \nonumber\\
		&\leq \frac{|\alpha|_\infty}{1-|\alpha|_\infty}\|_{\infty}f- S_{n,\rho}(f,\cdot)\|_{\infty}+\frac{1}{1-|\alpha|_\infty}\|S_{N,\rho}(f,\cdot)-f\|_{\infty}\nonumber\\
		&\leq  \frac{|\alpha|_\infty}{1-|\alpha|_\infty}\omega\left(f,\frac{b-a}{n}\right)+\frac{1}{1-|\alpha|_\infty}\omega\left(f, \frac{b-a}{N}\right).
	\end{align}\\
	
	\begin{theorem}\label{Thm5}
		For any function $f\in\mathcal{C}[a,b]$ and for every choice of scaling vector $\alpha=(\alpha_{1}, \alpha_{2}, \cdots, \alpha_{N})$ with $|\alpha_{i}|<1,~ \quad i =1,2,\cdots,N$, the double sequence $\left\{\left\{f_{n, N}^{\alpha}\right\}_{n=1}^{\infty}\right\}_{N=1}^{\infty}$ of neural network $\alpha$-fractal functions converges uniformly to the original function $f$ as $n, N\rightarrow \infty$.
	\end{theorem}
	
	\begin{proof}
		Proof of this theorem follows from Equation (\ref{omega}) by using the property that $\omega(f,\delta)\rightarrow0~ as~ \delta \rightarrow 0$. We have from Equation (\ref{omega}) that  $\frac{b-a}{n}\rightarrow 0~as~ n\rightarrow \infty$ and $\frac{b-a}{N}\rightarrow 0~as~ N \rightarrow \infty$. Thus, $\|f_{n, N}^{\alpha}-f\|\rightarrow 0~as~ n,N\rightarrow\infty$. 
	\end{proof} 
	\subsection{Numerical experiments}
	To validate the theoretical results obtained, we provide some numerical examples with graphs as: 
	\begin{example}\label{ex11}
		Let the target function be $f(x)=2\sin(2x)$ and the uniform partition $\Delta=[x_0, x_1, x_2, \cdots, x_{N}]$ be the partition of $[0,1]$, where $x_{i}=\frac{i}{N}$, $i=0, 1, \cdots, N$. Let the sigmoidal function $\rho$ used in the NNOP $\mathcal{S}_{n,\rho}(f)$ be same as given in \ref{a1}
		
		\begin{figure}[htb!]
			\begin{subfigure}{.5\textwidth} 
				\includegraphics[width=1\linewidth, height=5cm]{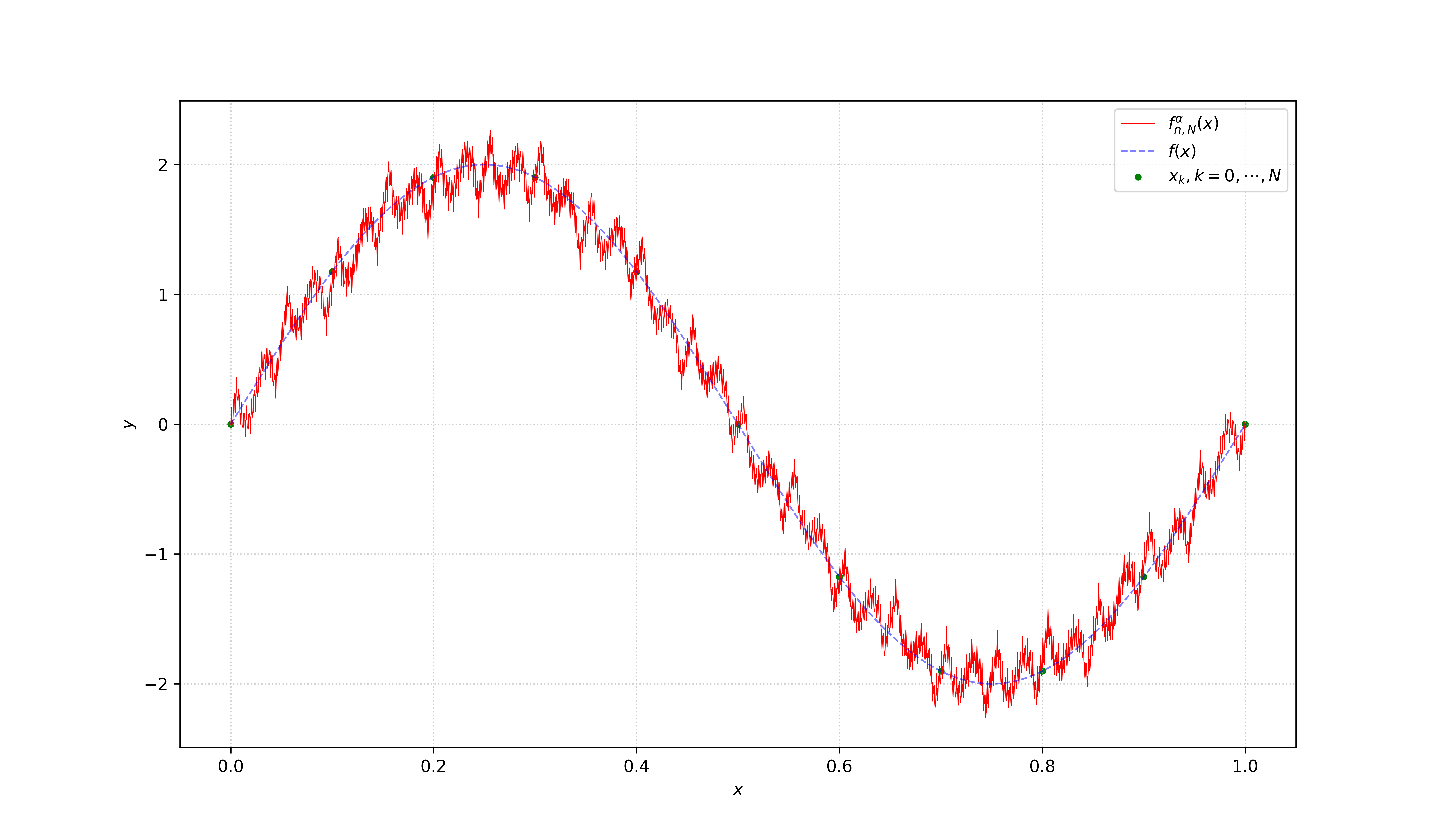} 
				\caption{$f_{5,11}^{0.5}$}
				\label{d1}
			\end{subfigure}
			\begin{subfigure}{.5\textwidth}
				\includegraphics[width=1\linewidth, height=5cm]{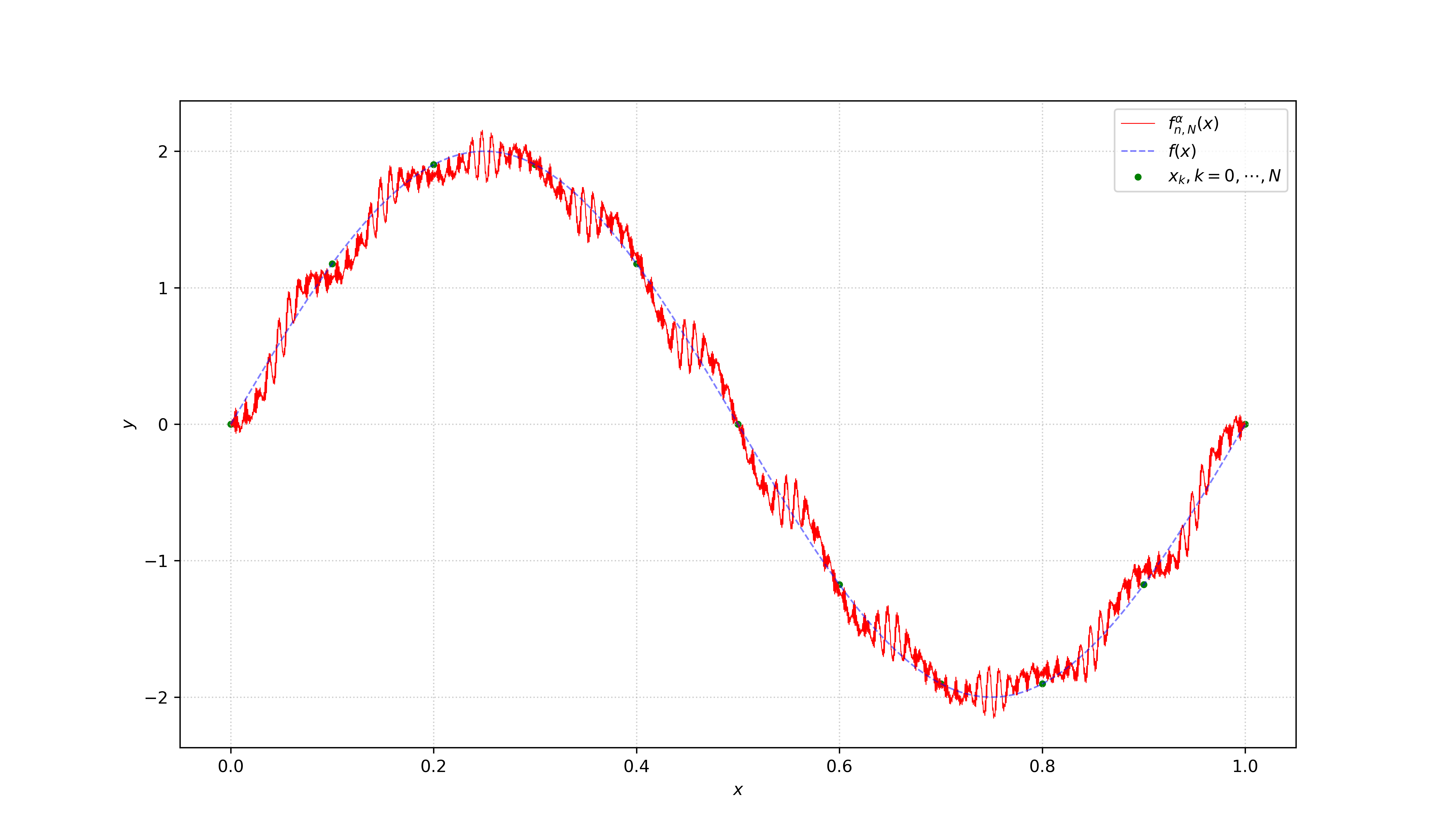}
				\caption{$f_{10,11}^{0.5}$}
				\label{d2}
			\end{subfigure}
			\begin{subfigure}{.5\textwidth}
				\includegraphics[width=1\linewidth, height=5cm]{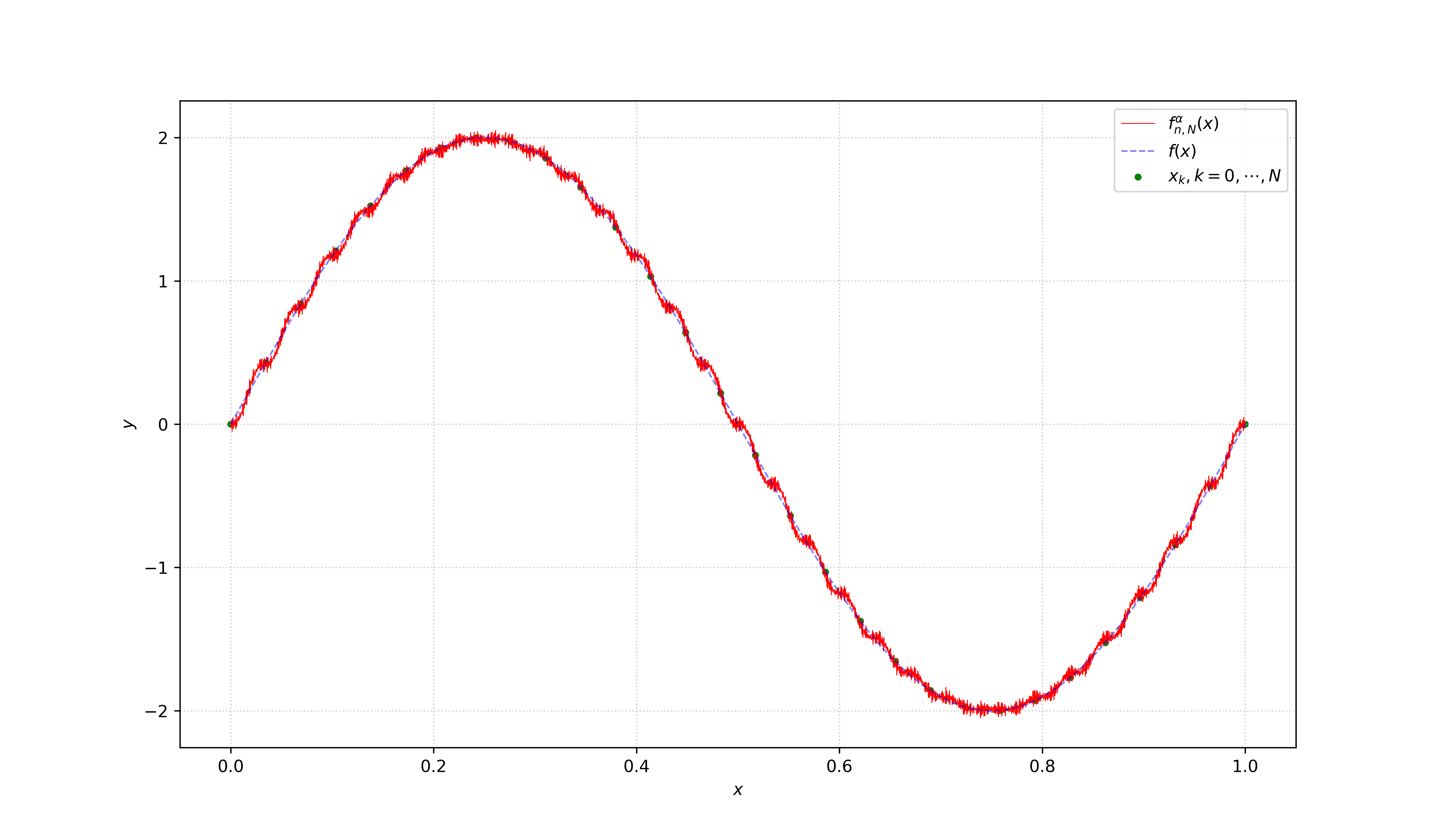}
				\caption{$f_{15,11}^{0.5}$}
				\label{d3}
			\end{subfigure}
			\begin{subfigure}{.5\textwidth}
				\includegraphics[width=1\linewidth, height=5cm]{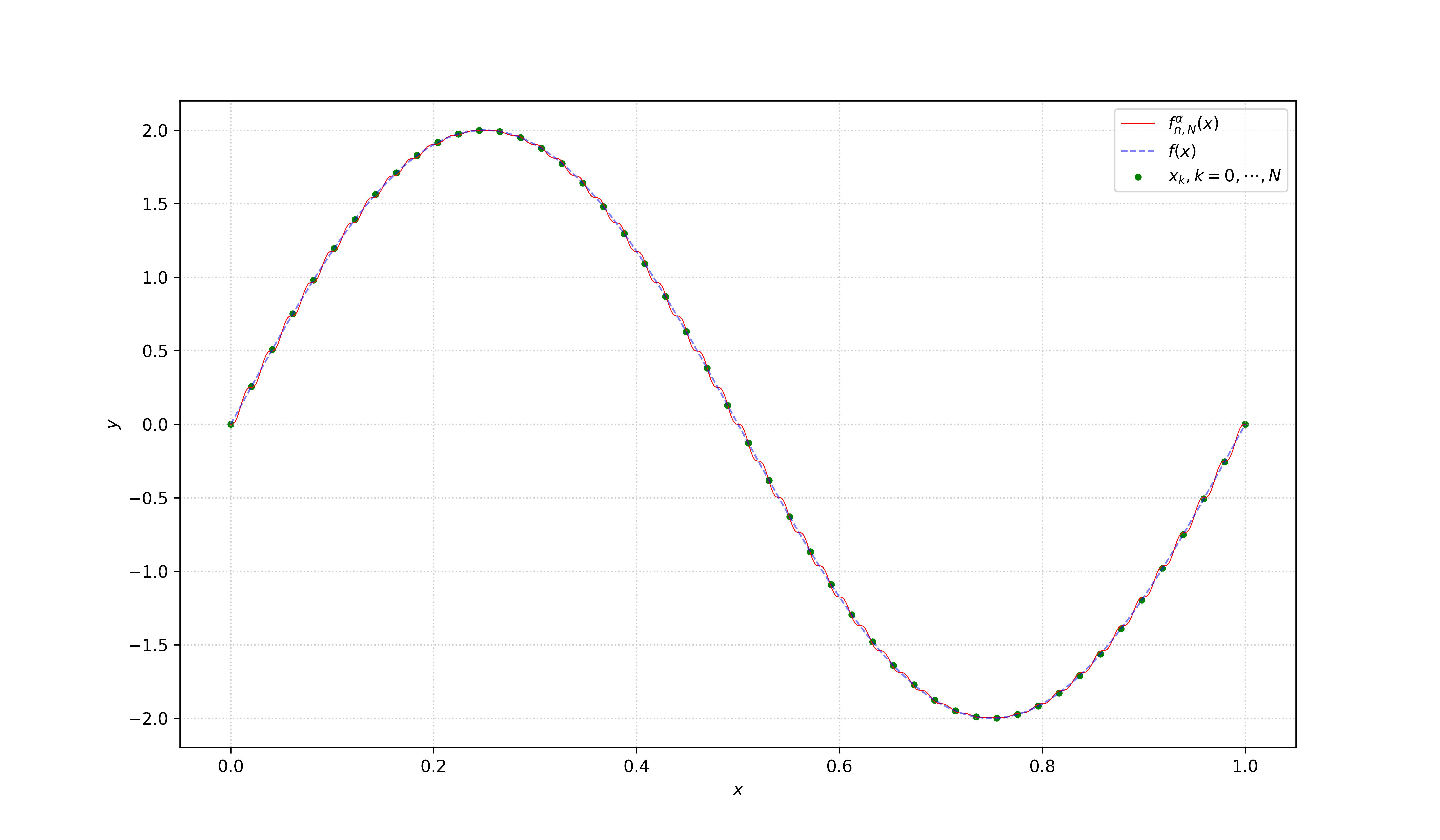}
				\caption{$f_{50,50}^{0.2}$}
				\label{d4}
			\end{subfigure}
			
			\caption{Comparison among fractal approximation for different values of $n$, $N$, and $\alpha$}\label{d}
		\end{figure}
		
		Figure \ref{d} (a-d) represents the fractal functions $f_{n, N}^{\alpha}$ for different values of $n$, $N$, and $\alpha$. Figure \ref{d1} represents  $f_{5, 11}^{0.5}$, while figures \ref{d2}, \ref{d3}, and \ref{d4} represents $f_{10, 11}^{0.5}$, $f_{40, 30}^{0.5}$, and $f_{50, 50}^{0.2}$ respectively. It can be seen that the fractal function $f_{n, N}^{\alpha}$ approaches to the original function with higher values of $n, N$ as the theoretical result obtained in Theorem \ref{Thm5}.
	\end{example}

	\section{Smooth fractal interpolation functions}
	
	In this section, we constructed smooth FIFs by exploring iterated function system (IFS) that satisfy the hypotheses of the theorem of Barnsley and Harrington \cite{BARNSLEY198914}. This result establishes the existence of differentiable FIFs and provides the necessary conditions for their differentiability. By employing a four-layered neural network operator, $\alpha$-fractal functions are constructed that preserve the smoothness of the target function, i.e., whenever $f \in C^r[a, b]$, the corresponding fractal function $f^\alpha \in C^r[a, b]$.\\

	\begin{theorem}\cite{BARNSLEY198914}  \label{brans}
		Let $\{a=x_0, x_1,x_2, \cdots, x_N=b\}$ be a partition of closed interval [a,b]. Denote by \(\pounds_i(x)\) the affine function satisfying conditions (\ref{ln}) and let \(a_i = \pounds_i^{\prime} = \frac{x_i - x_{i-1}}{x_N - x_0}\). Define \(\mathcal{J}_{i}(x, y) = \alpha_i y + \wp_{i}(x)\), for \(i = 1, 2, \ldots, N\). Let \(r \geqslant 0\) be a positive integer such that \(\left|\alpha_i\right| < a_i^r\) and \(\wp_{i} \in C^r\left[a, b\right]\) for all \(i = 1, 2, \ldots, N\). Suppose 
		\begin{align}  
			&\mathcal{J}_{i k}(x, y) = \frac{\alpha_i y + \wp_{i}^{(k)}(x)}{a_i^k}, \quad k=1,2,\cdots,r, \\  
			&y_{0, k} = \frac{\wp_1^{(k)}\left(x_0\right)}{a_1^k - \alpha_1}, \quad y_{N, k} = \frac{\wp_N^{(k)}\left(x_N\right)}{a_N^k - \alpha_N} \quad  k=1,2,\cdots,r .  
		\end{align}  
		If \(\mathcal{J}_{i-1, k}\left(x_N, y_{N, k}\right) = \mathcal{J}_{i k}\left(x_0, y_{0, k}\right)\) for \(i = 2, 3, \ldots, N\) and \(k = 1, 2, \ldots, n\), then the IFS 
		\begin{equation*}
			\left\{\left[a,b]\times \mathbf{C},~ (\pounds_i(x), \mathcal{J}_{i}(x, y)\right):~i=1,2,\cdots,N\right\}
		\end{equation*} 
		defines a fractal interpolation function \(f \in C^r\left[a,b\right]\). Furthermore, the derivatives \(f^{(k)}\) are fractal interpolation functions determined by 
		\begin{equation*}
			\left\{[a,b]\times \mathbf{C},~
			\pounds_i(x),~ \mathcal{J}_{ik}(x,y): ~ i =1,2,\cdots,N\right\}
		\end{equation*} for \(k = 1, 2, \ldots, n\).  
	\end{theorem}
	
	Consider a function $f\in C^{r}[a,b]$. Let $\Delta=\left\{x_0,x_1,\cdots, x_N\right \}$, where $x_k=a+kh,~~ k=0,1,2,\cdots,N \text{ and } h=\frac{b-a}{N}$ the uniform partition of interval $[a,b]$. Define the IFS through the functions:
	\begin{equation}\label{l22}
		\pounds_i(x)= \frac{x_i -x_{i-1}}{x_N-x_0}.x +\frac{x_N x_{i-1} -x_0 x_i}{x_N-x_0}, \quad \forall i=1,2,\cdots,N
	\end{equation}
	and 
	\begin{equation}\label{f22}
		\mathcal{J}_i(x,y)=a_{i}.y+ \wp_{i}(x), \quad \forall i=1,2,\cdots,N.
	\end{equation}
	Let
	\begin{equation} \label{qi1}
		\wp_{i}(x)= f(\pounds_{i}(x))-\alpha_{i}S_{n,r,\rho}(f, x)
	\end{equation}
	where $S_{n,r,\rho}(f, \cdot)$ is a neural network operator with four layers introduced by Qian et al in \cite{QIAN2022126781} and is given by
	\begin{equation*}
		S_{n,r,\rho}(f, x)=\sum_{j=0}^{r}\sum_{k=0}^{n}\Omega_{k,j}\Psi_{j}\left(\frac{2m}{h}\left(x-x_k\right)\right)
	\end{equation*}
	where $\Omega_{k,j}=\frac{h^j}{(2m)^j j!}f^{(j)}(x_k)$ and $\Psi_j\left(\frac{2m}{h}\left(x-x_k\right)\right) =\left(\frac{2m}{h}\left(x-x_k\right)\right)^{j}\psi\left(\frac{2m}{h}\left(x-x_k\right)\right).$
	$\psi$ is defined in Equation (\ref{xi})\\
	
	\begin{theorem}
		Let $f, \psi\in C^{r}[a,b]$ and  $\Delta=\left\{x_0,x_1,\cdots, x_N\right \}$, where $x_k=a+kh,~~ k=0,1,2,\cdots,N \text{ and } h=\frac{b-a}{N}$ be the uniform partition of interval $[a,b]$. Let $\pounds_i(x)$ and $\mathcal{J}_{i}(x,y)$ be defined by Equations (\ref{l22}) and (\ref{f22}). Choose $\alpha_{i}$ such that $|\alpha_{i}| <\frac{1}{N^r}$. Then the IFS $\Lambda=\left\{[a,b]\times \mathbf{C},
		\pounds_i(x),\mathcal{J}_{i}(x,y):~i =1,2,\cdots,N\right\}$ determines a fractal interpolation function, say $f^\alpha \in C^r[a,b]$ and the $k^{th}$ derivative is determined by IFS  $\mathcal{I}=\left\{[a,b]\times \mathbf{C},~
		\pounds_i(x),~ \mathcal{J}_{ik}(x,y): \quad i =1,2,\cdots,N\right\}$ where
		\begin{align*}  
			\mathcal{J}_{i k}(x, y) = \frac{\alpha_i y + \wp_{i}^{(k)}(x)}{a_i^k}, \quad k=1,2,\cdots,r. 
		\end{align*}  
	\end{theorem}
	\begin{proof}
		First we show that $\wp_{i}\in C^r[a,b]\quad \forall~ \quad i =1,2,\cdots,N$:\\
		Since $f \in C^r[a,b]$ and from \cite{QIAN2022126781} $\mathcal{S}_{n,r,\rho}(f,\cdot) \in C^r[a,b]$, implies $\wp_{i} \in C^r[a,b]$.\\
		Also from Equation (\ref{l22}), we have
		\begin{equation*}
			\pounds^{\prime}_i(x)= \frac{x_i -x_{i-1}}{x_N-x_0}=\frac{1}{N}
		\end{equation*} 
		Now our claim is that
		\begin{equation*}
			\mathcal{J}_{i-1, k}\left(x_N, y_{N, k}\right) = \mathcal{J}_{i k}\left(x_0, y_{0, k}\right)
		\end{equation*}
		where 
		\begin{align*}  
			&\mathcal{J}_{i k}(x, y) = \frac{\alpha_i y + \wp_{i}^{(k)}(x)}{a_i^k}, \quad k=1,2,\cdots,r, \\  
			&y_{0, k} = \frac{\wp_1^{(k)}\left(x_0\right)}{a_1^k - \alpha_1}, \quad y_{N, k} = \frac{\wp_N^{(k)}\left(x_N\right)}{a_N^k - \alpha_N} \quad  k=1,2,\cdots,r .  
		\end{align*} 
		Proof of claim:\\
		From Equation (\ref{qi}) and using the properties of $\pounds_i$ and $\mathcal{S}_{n,r.\rho}(f,\cdot)$, we have 
		\begin{align*}
			y_{0, k}=\frac{\frac{f^{k}(\pounds_1(x_0))}{N^k}-\alpha_{1}\mathcal{S}^{(k)}_{n,r.\rho}(f,x_0)}{\frac{1}{N^k}-\alpha_{1}}=f^k(x_0).
		\end{align*}
		Similarly, we can show that
		\begin{align*}
			y_{N, k}=f^k(x_N).
		\end{align*}
		Now, 
		\begin{align*}
			\mathcal{J}_{i-1, k}\left(x_N, y_{N, k}\right) &= \frac{\alpha_{i-1} y_{N,k} + (f(\pounds_{i-1}(x)))_{x=x_N}^{(k)}-\alpha_{i-1}S^{(k)}_{n,r,\rho}(f, x)_{(x=x_N)}}{\frac{1}{N^k}}\\
			&=\frac{\alpha_{i-1} f^k(x_N) + \frac{f^k(x_{i-1})}{N^k}-\alpha_{i-1}f^k(x_N)}{\frac{1}{N^k}}\\
			&=f^k(x_{i-1})
		\end{align*}
		and
		\begin{align*}
			\mathcal{J}_{i, k}\left(x_0, y_{0, k}\right) &= \frac{\alpha_{i} y_{0,k} + (f(\pounds_{i}(x)))_{x=x_0}^{(k)}-\alpha_{i}S^{(k)}_{n,r,\rho}(f, x)_{(x=x_0)}}{\frac{1}{N^k}}\\
			&=\frac{\alpha_{i} f^k(x_0) + \frac{f^k(x_{i-1})}{N^k}-\alpha_{i}f^k(x_0)}{\frac{1}{N^k}}\\
			&=f^k(x_{i-1}),
		\end{align*}
		which proves our claim.
		\\
		Therefore, our theorem satisfies the hypothesis of Barnsley and Harrington Theorem \ref{brans}, and hence the proof follows.
		
	\end{proof}
	
	\section{Conclusion}\label{sec13}
	
	This article introduces a new approach to constructing fractal interpolation functions (FIFs) using neural network operators. We demonstrated that the newly constructed $\alpha$-fractal functions can approximate continuous functions to any preassigned degree of accuracy and established uniform convergence results, showing that the $\alpha$-fractal functions converge to the original function for every choice of the scaling parameter $\alpha$ with the help of the modulus of continuity.   By employing a four-layered neural network operator, we have also constructed $\alpha$-fractal functions that preserve the smoothness of the target function. The key contribution is that, in Section \ref{sec5}, we have developed fractal interpolation functions that utilize only the values of the original function at the nodes or partition points, unlike traditional methods that depend on each function value. This discrete approximation method reduces the computational cost while maintaining the essential features of the target function. With the help of the modulus of continuity, we have obtained uniform error estimates. This work bridges the gap between neural network operators and fractal interpolation, offering a powerful tool in approximation theory.
	
	\section*{Declarations}

	\begin{itemize}
		\item Funding: Not available
		\item Conflict of interest: The authors declare that they have no competing interests.
		\item Data availability:
		Not applicable 
		
	\end{itemize}



\begin{thebibliography}{10}
			
			\bibitem{nasim}
			M.~N. Akhtar, M.~G.~P. Prasad, and M.~A. Navascu\'{e}s.
			\newblock Box dimensions of $\alpha$-fractal functions.
			\newblock {\em Fractals}, 24(03):1650037, 2016.
			
			\bibitem{Anastassiou2000}
			G.~A. Anastassiou and S.~G. Gal.
			\newblock {\em Approximation Theory}.
			\newblock Birkh\"{a}user Boston, 2000.
			
			\bibitem{articlekj}
			K.~Ansari, S.~Karakilic, and F.~Özger.
			\newblock Bivariate {B}ernstein-{K}antorovich operators with a summability
			method and related {GBS} operators.
			\newblock {\em Filomat}, 36, 12 2022.
			
			\bibitem{aymanfilomat} M.~ Ayman-Mursaleen, B.~P. Lamichhane, A. Kiliçman, and  N. Senu  On $q$-statistical approximation of wavelets aided Kantorovich $q$-Baskakov operators. {\em Filomat}, 38(9), 3261–3274, 2024.
			
			\bibitem{barnsley1986fractal}
			M.~F. Barnsley.
			\newblock Fractal functions and interpolation.
			\newblock {\em Constructive approximation}, 2:303--329, 1986.
			
			\bibitem{BARNSLEY198914}
			M.~F. Barnsley and A.~N. Harrington.
			\newblock The calculus of fractal interpolation functions.
			\newblock {\em J. Approximation Theory}, 57(1):14--34, 1989.
			\bibitem{BHAT2026190}
			A.~A. Bhat and A.~Khan.
			\newblock Boundary interpolation on triangles via neural network operators.
			\newblock {\em Math. Comput. Simul}, 241:190--201, 2026.
			
			\bibitem{bhat2026blending}
			A.~A. Bhat, A.~Khan, M.~Iliyas, K.~Khan, and M.~Mursaleen.
			\newblock A blending approach to transfinite interpolation on compact disks using neural network operators.
			\newblock {\em Iran. J. Sci.}, Nov. 2025.
			
			\bibitem{Brewal2026}
			S. ~Berwal, , S.~A. Mohiuddine, A.~Kajla., and  A.~ Alotaibi, \newblock Theoretical and Numerical Approximation Through Generalized Fractional Integral Type Max Product Neural Network Operators.
			\newblock  {\em Numer. Funct. Anal. Optim.}, 47(1-6), 3-29. 2026
			
			\bibitem{cai2023approximation}
			Q.~B. Cai, A.~Khan, M.~S. Mansoori, M.~Iliyas, and K.~Khan.
			\newblock Approximation by $\lambda$-{B}ernstein type operators on triangular
			domain.
			\newblock {\em Filomat}, 37(6):1941--1958, 2023.
			
			\bibitem{cai2018shape}
			Q.~B. Cai and G. Zhou 
			\newblock Blending type approximation by {GBS} operators of bivariate tensor
			product of $\lambda$-{B}ernstein--{K}antorovich type.
			\newblock {\em J. Inequal. Appl.}, 2018(1) 268, 2018.
			
			\bibitem{COSTARELLI201528}
			D.~Costarelli.
			\newblock Neural network operators: Constructive interpolation of multivariate
			functions.
			\newblock {\em Neural Networks}, 67:28--36, 2015.
			
			\bibitem{COSTARELLI201480}
			D.~Costarelli and R.~Spigler.
			\newblock Convergence of a family of neural network operators of the
			{K}antorovich type.
			\newblock {\em J. Approximation Theory}, 185:80--90, 2014.
			
			\bibitem{HORNIK1989359}
			K.~Hornik, M.~Stinchcombe, and H.~White.
			\newblock Multilayer feedforward networks are universal approximators.
			\newblock {\em Neural Networks}, 2(5):359--366, 1989.
			
			\bibitem{husain2024novel}
			A.~Husain, S.~Gumma, M.~Sajid, J.~Reddy, and M.~T. Alresheedi.
			\newblock A novel fractal interpolation function algorithm for fractal
			dimension estimation and coastline geometry reconstruction: a case study of
			the coastline of {K}ingdom of {S}audi {A}rabia.
			\newblock {\em The European Physical Journal B}, 97(4):51, 2024.
			
			\bibitem{husain2021fractal}
			A.~Husain, J.~Reddy, D.~Bisht, and M.~Sajid.
			\newblock Fractal dimension of coastline of {A}ustralia.
			\newblock {\em Sci. Rep.}, 11(1):6304, 2021.
			
			\bibitem{khan2021phillips}
			A.~Khan, M.~Mansoori, K.~Khan, and M.~Mursaleen.
			\newblock Phillips-type q-{B}ernstein operators on triangles.
			\newblock {\em J. Funct. Spaces	}, 2021(1):6637893, 2021.
			
			\bibitem{doi:10.1126/science.156.3775.636}
			B.~Mandelbrot.
			\newblock How long is the coast of {B}ritain? statistical self-similarity and
			fractional dimension.
			\newblock {\em Science}, 156(3775):636--638, 1967.
			
			
			\bibitem{MASSOPUST1997171}
			P.~R. Massopust.
			\newblock Fractal functions and their applications.
			\newblock {\em Chaos, Solitons Fractals}, 8(2):171--190, 1997.
			
			
			\bibitem{MURSALEEN2015874}
			M. Mursaleen, K. J. Ansari, A. Khan, On $(p,q)$%
			-analogue of Bernstein Operators.
			 \newblock {\em Appl. Math. Comput.}, 266 (2015)
			874-882, [Erratum: (2016), 278, 70-71].
			
			\bibitem{vijenderbernstein}
			V.~ Nallapu.
			\newblock Bernstein fractal approximation and fractal full Müntz theorems.
			\newblock {\em Electron. Trans. Numer. Anal.}, 51:1--14, 2019.
			
			\bibitem{Navascues2005}
			M.~Navascues.
			\newblock Fractal trigonometric approximation.
			\newblock {\em Electron. Trans. Numer. Anal.}, 20:64--74, 2005.
			
			\bibitem{navascues2010fractal}
			M.~Navascu{\'e}s.
			\newblock Fractal approximation.
			\newblock {\em Complex Anal. Oper. Theory}, 4:953--974, 2010.
			
			
			\bibitem{navascues2006smooth}
			M.~Navascu{\'e}s and M.~Sebasti{\'a}n.
			\newblock Smooth fractal interpolation.
			\newblock {\em J. Inequal. Appl.}, 2006:1--20, 2006.
			
			\bibitem{QIAN2022126781}
			Y.~Qian and D.~Yu.
			\newblock Rates of approximation by neural network interpolation operators.
			\newblock {\em Appl. Math. Comput.}, 418:126781, 2022.
			
			\bibitem{sajid2023box}
			M.~Sajid, A.~Husain, J.~Reddy, M.~T. Alresheedi, S.~A. Al~Yahya, and
			A.~Al-Rajy.
			\newblock Box dimension of the border of {K}ingdom of {S}audi {A}rabia.
			\newblock {\em Heliyon}, 9(4), 2023.
			
			\bibitem{far}
			F.~Özger.
			\newblock Applications of generalized weighted statistical convergence to
			approximation theorems for functions of one and two variables.
			\newblock {\em Numer. Funct. Anal. Optim.},
			41(16):1990--2006, 2020.
			
		\end{thebibliography}

	\end{document}